\documentclass[11pt]{article}

\usepackage[preprint]{acl}
\usepackage{booktabs}
\usepackage{times}
\usepackage{latexsym}
\usepackage[most]{tcolorbox}
\usepackage{tabularx}

\usepackage[T1]{fontenc}

\usepackage[utf8]{inputenc}
\usepackage{microtype}

\usepackage{inconsolata}
\usepackage{subcaption}

\usepackage{graphicx}

\usepackage{placeins}

\usepackage{amsmath} 
\usepackage{paralist}

\usepackage{xcolor}
\usepackage{framed}
\usepackage{enumitem}

\title{
Calibrating LLM Judges: Linear Probes for Fast and Reliable Uncertainty Estimation}

\author{
 \textbf{Bhaktipriya Radharapu\textsuperscript{1}},
 \textbf{Eshika Saxena\textsuperscript{1}},
 \textbf{Kenneth Li\textsuperscript{2}},
 \textbf{Chenxi Whitehouse\textsuperscript{1}},
\\
 \textbf{Adina Williams\textsuperscript{1}},
 \textbf{Nicola Cancedda\textsuperscript{1}}
\\
 \textsuperscript{1}FAIR at Meta,
  \textsuperscript{2}Meta Superintelligence Labs
\\
 \small{
   \textbf{Correspondence:} \href{mailto:email@domain}{bhakti@meta.com}
 }
}

\begin{document}
\maketitle

\begin{abstract}
As LLM-based judges become integral to industry applications, obtaining well-calibrated uncertainty estimates efficiently has become critical for production deployment. However, existing techniques, such as verbalized confidence and multi-generation methods, are often either poorly calibrated or computationally expensive. We introduce linear probes trained with a Brier score-based loss to provide calibrated uncertainty estimates from reasoning judges' hidden states, requiring no additional model training. We evaluate our approach on both objective tasks (reasoning, mathematics, factuality, coding) and subjective human preference judgments. Our results demonstrate that probes achieve superior calibration compared to existing methods with $\approx10$x computational savings, generalize robustly to unseen evaluation domains, and deliver higher accuracy on high-confidence predictions. However, probes produce conservative estimates that underperform on easier datasets but may benefit safety-critical deployments prioritizing low false-positive rates. Overall, our work demonstrates that interpretability-based uncertainty estimation provides a practical and scalable plug-and-play solution for LLM judges in production.

\end{abstract}

\section{Introduction}

\begin{figure}
    \centering
    \includegraphics[width=\linewidth]{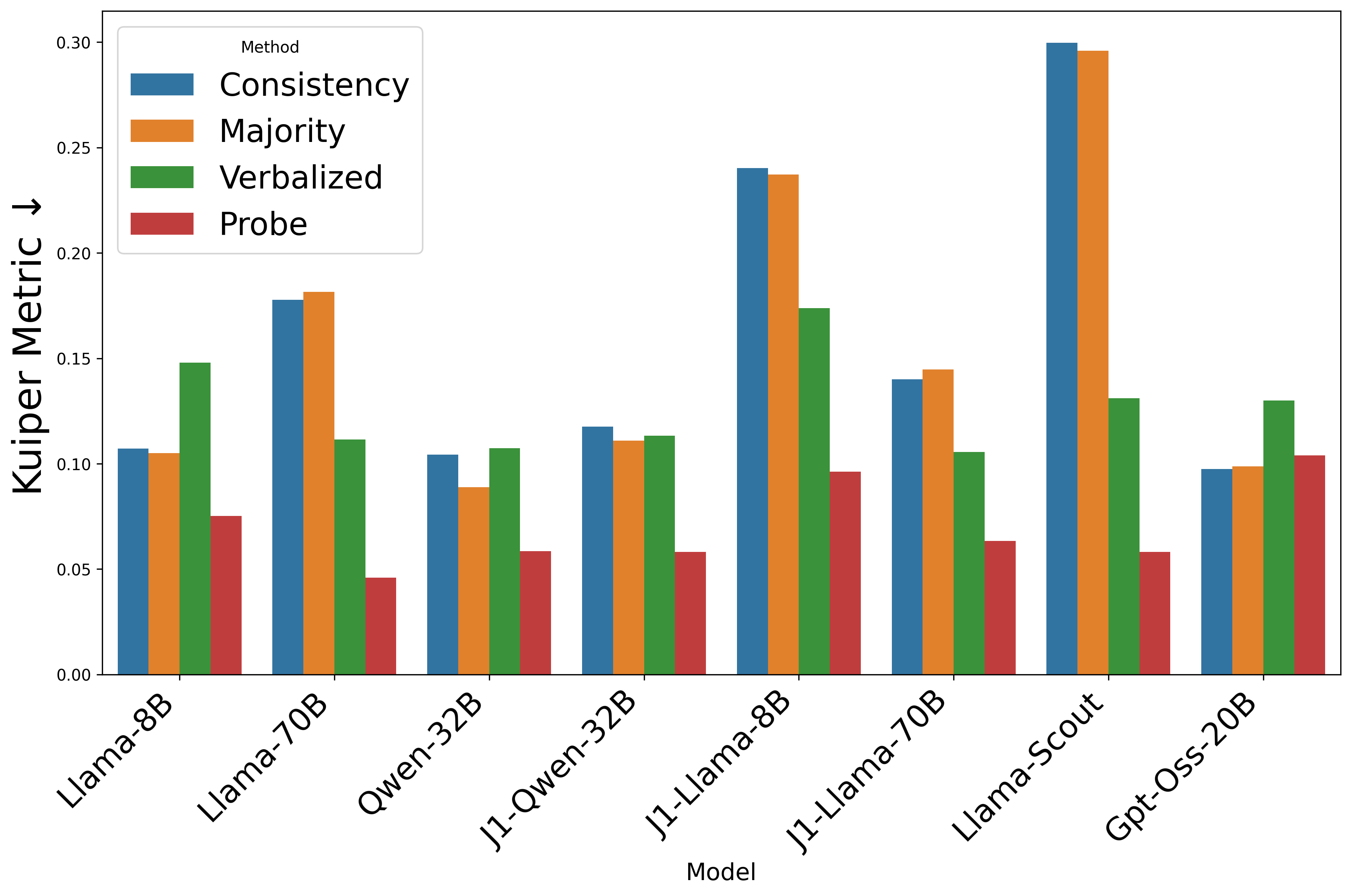}
            \caption{Calibration of models across model architectures, datasets, and uncertainty estimation methods as measured by the Kuiper metric. We compare four approaches across dense prompted judges (LLaMA 8B/70B, Qwen 32B), dense fine-tuned judges (J1 family), and MoE prompted judges (GPT-OSS 20B, LLaMA Scout). Our probe-based method outperforms baseline approaches across all architectures and training paradigms. Results averaged across all evaluation datasets. \textbf{Lower values indicate better calibration.}}

    \label{fig:my_label}
\end{figure}
LLM-as-judge paradigm \citep{zheng2023judging, dubois2023alpacafarm, touvron2023llama2openfoundation} has become ubiquitous in modern AI development, providing reward signals for alignment training (RLAIF) \citep{bai2022constitutional, lee2023rlaif, kirk2024prism} and ranking models for deployment decisions \citep{chiang2024chatbot}. With the rapid advancement of reasoning-capable models \citep{openai2024o1, deepseek2025r1}, LLM judges that generate reasoning traces before rendering verdicts are becoming increasingly prevalent due to their higher accuracy and transparency \citep{whitehouse2025j10, chen2025judgelrmlargereasoningmodels}.

Yet, current practice treats all judgments as equally reliable. Without calibrated confidence estimates, we cannot distinguish high-confidence judgments from cases where the LLM is essentially guessing. Moreover, LLM judges are known to be overconfident, systematically expressing higher confidence than their empirical accuracy supports \citep{tian2025overconfidence, jung2024trust}. This leaves practitioners to either blindly trust all judgments or manually review everything (defeating the purpose of automation). This lack of uncertainty awareness is particularly problematic across judgment types: for correctness evaluation, we cannot identify objectively wrong judgments to exclude \citep{hongli-etal-2024-mitigating, ross2024what, wang2023large}; for preference evaluation, we cannot detect genuinely ambiguous cases where humans may disagree or multiple valid answers exist \citep{Aroyo_Welty_2015, pavlick-kwiatkowski-2019-inherent, nie-etal-2020-learn, talat-etal-2022-machine, radharapu2025arbiters}.

Calibrated LLM judges, whose expressed confidence matches their empirical accuracy, enable systems to route straightforward cases to efficient models while reserving expensive judges or human reviewers for uncertain decisions \citep{jung2024trust, chen2023frugalgpt}, dramatically reducing computational and labor costs. Training processes become more efficient by down-weighting uncertain judgments, preventing noisy labels from causing reward hacking \citep{gao2023scaling} and model collapse \citep{zhang2024diverging}. Across these applications, calibration serves complementary purposes: flagging likely errors in correctness tasks while preserving valid diversity in preference tasks. Calibration is thus essential for building reliable and trustworthy AI systems with LLM judges.

In this work, our main contributions are: (1) A Brier-score–trained linear probe that produces calibrated uncertainty estimates from reasoning judges' hidden states, requiring no additional model training or multi-sample generation. (2) Our probe achieves substantially better calibration than verbalized and multi-generation baselines across multiple model families (dense and MoE) and judging styles (prompted and fine-tuned), while requiring an order of magnitude less compute. (3) Strong out-of-distribution generalization to unseen benchmarks, with analysis of key trade-offs in probe's performance.

\section{Related Work}

\subsection{LLM Judges and Reasoning-Based Evaluation}

Large language model judges have become ubiquitous for evaluating model outputs across tasks ranging from pairwise preference ranking for RLHF~\citep{christiano2017deep, ouyang2022training} to judging for correctness in verifiable tasks. These judges span a spectrum from prompted general-purpose models like GPT-4~\citep{zheng2023judging, dubois2023alpacafarm} to specialized fine-tuned judges \citep{whitehouse2025j10, zhu2023judgelm, kim2024prometheus, kim2024prometheus2, li2024generative} that reason before giving their verdict.  While these judges achieve high accuracy, LLM judges are known to suffer from systematic overconfidence~\citep{jung2024trust, tian2025overconfidence, xiong2024can}.

\subsection{Uncertainty Estimation in Large Language Models}

Uncertainty estimation methods for LLMs fall into several categories, each with distinct trade-offs:

\paragraph{Logit-Based Methods.} Token-level uncertainty estimation methods like Perplexity and Temperature Scaling~\citep{guo2017calibration} assume uniform calibration across tokens, ignoring the semantic and contextual nuances crucial for reasoning tasks~\citep{xie2024calibrating}. Maximum Softmax Probability (MSP)~\citep{plaut2024probabilities} has been shown to be consistently miscalibrated in reasoning and multiple-choice QA tasks. Additionally, approaches such as contextual calibration~\citep{zhao2021calibrate} and batch calibration~\citep{zhou2024batch} operate on single-token output logits, making them inapplicable to reasoning judges that generate multi-token responses.

\paragraph{Verbalized Confidence.} Asking models directly for confidence scores~\citep{tian2023just, kadavath2022language, lin2023teaching} is straightforward, and has been argued to be better than logit-based methods. However, verbalized confidence produces overconfident estimates ~\citep{tian2025overconfidence,xiong2024can, tao2025revisiting, lyu2025calibrating}. 

\paragraph{Consistency-Based Methods.} Self-consistency \citep{wang2022self0consistency, manakul2023selfcheckgpt}, semantic entropy~\citep{kuhn2023semantic}, and related approaches~\citep{chen-mueller-2024-quantifying} achieve strong calibration by aggregating uncertainty across multiple generations. Methods such as prompt ensembles or model ensembles \cite{huang2023look, jung2024trust, tian2025overconfidence} involve perturbing prompts or aggregating uncertainty from different models. However, they incur substantial computational costs (typically 10--20$\times$ inference overhead), limiting practical deployment.

\paragraph{Training-Based Approaches.} Methods like fine-tuning the model for improved verbalized confidence ~\citep{damani2025beyond, li2025conftuner0, li2025judging} or specialized architectures ~\citep{huang2023look, kapoor-etal-2024-calibration, xie2024calibrating} require significant computational overhead through additional training or architectural modifications.

\paragraph{Interpretability-Based Approaches.} 
Recent work has extracted uncertainty signals from internal representations~\citep{azaria2023internal, zou2023representation, li2024inference, kossen2024semantic, burns2023discovering,sriramanan2024llmcheck}, but focuses predominantly on hallucination detection in factual question-answering where models recall memorized knowledge—a fundamentally different setting from reasoning. Other methods~\cite{zhang2025reasoning, bi2025cot0kinetics0, wang2024latent, ji2025calibrating, NEURIPS2024_d037fd02} targetted at reasoning, store hidden states per layer and per intermediate reasoning step/token during inference, incurring significant computational overhead with longer reasoning chains. Moreover, these methods target \emph{uncertainty-based selective classification} (distinguishing correct from incorrect predictions) rather than \emph{uncertainty calibration} (aligning predicted uncertainty with actual correctness). These tasks are complementary, and performance on one does not imply the other~\citep{tao2025revisiting}.

Our work addresses these limitations with a plug-and-play approach: linear probes that achieve strong calibration for reasoning judges without requiring additional model training, multi-sample generation, or per-token state storage during inference, offering a practical and computationally efficient solution for deployment at scale.

\section{Method}

\begin{table*}
  \centering
    \footnotesize

  \begin{tabular*}{\textwidth}{@{\extracolsep{\fill}}l*{6}{c}@{}}
    \hline
    \textbf{Method} & \textbf{LLAMA 8B} & \textbf{LLAMA 70B} & \textbf{QWEN 32B} & \textbf{J1 QWEN 32B} & \textbf{J1 LLAMA 8B} & \textbf{J1 LLAMA 70B} \\
    \hline
    \multicolumn{7}{c}{\textbf{PPE Correctness}} \\
    \hline
    Verbalized & 0.182 / 0.174 & 0.135 / 0.131 & 0.107 / 0.108 & 0.103 / 0.105 & 0.219 / 0.213 & 0.076 / 0.074 \\
    Consistency & 0.159 / 0.170 & 0.211 / 0.208 & 0.132 / 0.145 & 0.098 / 0.130 & 0.278 / 0.260 & 0.134 / 0.144 \\
    Majority & 0.159 / 0.154 & 0.215 / 0.202 & 0.125 / 0.137 & 0.085 / 0.109 & 0.279 / 0.261 & 0.138 / 0.128 \\
    Probe & \textbf{0.047} / \textbf{0.075} & \textbf{0.017} / \textbf{0.025} & \textbf{0.020} / \textbf{0.028} & \textbf{0.029} / \textbf{0.039} & \textbf{0.047} / \textbf{0.069} & \textbf{0.017} / \textbf{0.018} \\
    \hline
    \multicolumn{7}{c}{\textbf{PPE Preference}} \\
    \hline
    Verbalized & 0.173 / 0.166 & 0.123 / 0.121 & 0.196 / 0.194 & 0.195 / 0.195 & 0.192 / 0.188 & 0.122 / 0.120 \\
    Consistency & 0.153 / 0.155 & 0.231 / 0.226 & 0.048 / 0.083 & 0.051 / 0.071 & 0.324 / 0.311 & 0.207 / 0.200 \\
    Majority & 0.150 / 0.143 & 0.234 / 0.223 & \textbf{0.038} / \textbf{0.065} & \textbf{0.050} / \textbf{0.069} & 0.323 / 0.310 & 0.211 / 0.197 \\
    Probe & \textbf{0.062} / \textbf{0.093} & \textbf{0.040} / \textbf{0.047} & 0.066 / 0.077 & 0.056 / 0.072 & \textbf{0.051} / \textbf{0.066} & \textbf{0.034} / \textbf{0.039} \\
    \hline
    \end{tabular*}
  \caption{Calibration on Preference Policy Evaluation (PPE) datasets measured by Kuiper/ECE (lower is better). We evaluate dense-architecture models as prompted and fine-tuned judges across multiple sizes. On objective PPE-Correctness tasks, probes outperform all baselines across all models. On subjective PPE-Preference tasks, probes achieve best performance on Llama variants and on Qwen closely match majority voting, which requires 10× higher computational cost.}
  \label{tab:iid_performance}
\end{table*}
\begin{table*}
  \centering
  \footnotesize
  \begin{tabular*}{\textwidth}{@{\extracolsep{\fill}}l*{6}{c}@{}}
    \hline
    \textbf{Method} & \textbf{LLAMA 8B} & \textbf{LLAMA 70B} & \textbf{QWEN 32B} & \textbf{J1 QWEN 32B} & \textbf{J1 LLAMA 8B} & \textbf{J1 LLAMA 70B} \\
    \hline
    \multicolumn{7}{c}{\textbf{JudgeBench}} \\
    \hline
    Verbalized & 0.205 / 0.188 & 0.152 / 0.145 & 0.103 / 0.107 & 0.114 / 0.113 & 0.274 / 0.263 & 0.156 / 0.155 \\
    Consistency & 0.077 / 0.193 & 0.231 / 0.243 & 0.205 / 0.276 & 0.087 / 0.124 & 0.282 / 0.259 & 0.188 / 0.204 \\
    Majority & 0.076 / 0.171 & 0.238 / 0.224 & 0.159 / 0.198 & 0.076 / 0.105 & 0.271 / 0.259 & 0.200 / 0.182 \\
    Probe & \textbf{0.072} / \textbf{0.119} & \textbf{0.062} / \textbf{0.077} & \textbf{0.055} / \textbf{0.059} & \textbf{0.037} / \textbf{0.048} & \textbf{0.122} / \textbf{0.135} & \textbf{0.075} / \textbf{0.074} \\
    \hline
    \multicolumn{7}{c}{\textbf{Reward Bench}} \\
    \hline
    Verbalized & \textbf{0.032} / \textbf{0.038} & \textbf{0.035} / 0.045 & \textbf{0.023} / \textbf{0.027} & \textbf{0.041} / \textbf{0.043} & \textbf{0.009} / \textbf{0.007} & 0.067 / 0.078 \\
    Consistency & 0.039 / 0.071 & 0.039 / 0.052 & 0.033 / 0.035 & 0.234 / 0.251 & 0.078 / 0.078 & \textbf{0.031} / 0.033 \\
    Majority & 0.035 / 0.059 & 0.040 / \textbf{0.041} & 0.034 / 0.035 & 0.232 / 0.249 & 0.076 / 0.075 & \textbf{0.031} / \textbf{0.031} \\
    Probe & 0.120 / 0.145 & 0.065 / 0.067 & 0.092 / 0.106 & 0.111 / 0.128 & 0.166 / 0.190 & 0.128 / 0.140 \\
    \hline
  \end{tabular*}
  \caption{Out-of-distribution calibration (Kuiper/ECE, lower is better). Probes outperform baselines on JudgeBench but lag on RewardBench, where higher accuracy makes verbalized method's overconfidence appear well-calibrated.}
\label{tab:ood_performance}
\end{table*}

Taking inspiration from other interpretability works on linear probes that by and large perform \emph{selective classification}
(distinguishing correct from incorrect predictions) \citep{kossen2024semantic, azaria2023internal, zou2023representation, burns2023discovering, li2024inference}, we train probes to improve \emph{calibration} of models.
We train a linear regression model on the judge's residual stream activations from the last token, with verdict accuracy as the label. Layer selection is based on validation performance, and we optimize the Brier score~\citep{glenn1950verification} loss:
$$\mathcal{L}_{Brier} = \frac{1}{N} \sum_{i=1}^{N} (\hat{y}_i - y_i)^2$$

where $N$ is the number of samples, $\hat{y}_i$ is the predicted probability of verdict accuracy, and $y_i \in \{0, 1\}$ is the ground truth label. See \ref{training_probe} for more details on the training of our probes.

In all our experiments, we prompt the judge to verbally express its uncertainty before its verdict, allowing the probe to leverage both internal representations and linguistic hedging signals, which correlate with uncertainty~\citep{lin2023teaching, lyu2025calibrating}.

\paragraph{Datasets.} We train our probes using the Preference Proxy Evaluation (PPE) dataset \citep{frick2024evaluate}, which features real-world prompts sourced from LM Arena and addresses two key LLM judge tasks: preference alignment and correctness. The dataset includes PPE Preference (10.2K samples) with human preference pairs from 20 LLMs across 121 languages, and PPE Correctness (12.7K samples) with response pairs from four models on five benchmarks (MMLU-Pro, MATH, GPQA, MBPP-Plus, IFEval) spanning knowledge, mathematics, STEM, coding, and instruction following.

For out-of-distribution evaluation, we use JudgeBench~\citep{tan2024judgebench0} (620 samples) and RewardBench~\citep{lambert2024rewardbench0} (3K samples). JudgeBench contains correct-incorrect response pairs across knowledge, reasoning, math, and coding tasks, while RewardBench focuses heavily on chat and human preference, and also includes safety, coding and reasoning. These datasets test our probes' ability to generalize to new prompt types, languages, tasks, and model responses beyond the training distribution.

\paragraph{Models.} We evaluate our probes on six dense models with varying architectures, model families, and training strategies: fine-tuned judges from the state-of-the-art J1 family \cite{whitehouse2025j10}, as well as their prompted (non-finetuned) variants across 8B, 32B, and 70B parameters---based on \citep{qwen3technicalreport} and Llama \citep{llama3modelcard}. In addition, we assess two Mixture-of-Experts (MoE) models: Llama-Scout (109B) \cite{meta_llama4_scout_17b_16e_instruct_2025} and GPT-OSS-20B \citep{openai2025gptoss120bgptoss20bmodel}. MoE models \citep{fedus2021switch} utilize sparse activation patterns and routing mechanisms, which may encode uncertainty differently than dense models, thereby providing a rigorous test of the robustness of our probing approach.

\paragraph{Judge Formulations.}
We consider three judge formulations: \textbf{Pairwise Judge with Verdict (PaV):} Given a question $x$ and response pair $(a, b)$, the judge generates reasoning trace followed by the preferred response $y$. \textbf{Pairwise Judge with Scores (PaS):} The judge generates reasoning trace, followed by real-valued scores $s_a, s_b$ for each response, selecting the higher-scoring response as the verdict. \textbf{Pairwise Judge with Likert Scale (PaL):} After generating the reasoning trace, the judge selects from \{$A \gg B$, $A > B$, Tie, $B > A$, $B \gg A$\}, where correctness is verified by checking if the winning response is ranked higher. We use PaV for Llama judges, PaS for Qwen judges (to maintain parity with the fine-tuned variants in J1 family \cite{whitehouse2025j10}), and PaL for GPT judges, following LM Arena hard prompt protocol \cite{tan2024judgebench0,li2024crowdsourced} (prompts in Appendix~\ref{prompt:likert}).

\paragraph{Evaluation metrics.} We evaluate probe performance using the Kuiper statistic and Expected Calibration Error (ECE). We use Kuiper statistic as the main indicator of calibration as it is the most stable of the two metrics. 
We train and evaluate probes on three different train-test splits of the PPE datasets, reporting average metrics across splits. Standard deviations were near zero for all results.

\paragraph{ECE (Expected Calibration Error).} ECE \citep{guo2017calibration, naeini2015obtaining} quantifies the difference between predicted confidence and actual accuracy by partitioning predictions into bins and measuring the gap within each bin. It is computed as: $$\text{ECE} = \sum_{m=1}^{M} \frac{|B_m|}{n} |\text{acc}(B_m) - \text{conf}(B_m)|$$ where $M$ is the number of bins, $B_m$ is the set of samples in bin $m$, $n$ is the total number of samples, $\text{acc}(B_m)$ is the accuracy in bin $m$, and $\text{conf}(B_m)$ is the average confidence in bin $m$. In our evaluation, we use a bin size of 0.1, as is common in the literature \cite{guo2017calibration,tao2025revisiting}, to partition the 0–1 interval. However, it is a known weakness that ECE is sensitive to bin size \citep{roelofs2020mitigating, nixon2019measuring}.

\paragraph{Kuiper statistic.} The Kuiper metric~\cite{tygert_2025_15253140, arrieta2022metrics} measures maximum cumulative miscalibration by computing the spread between over- and under-confidence across all thresholds. Given sorted scores $S_1 < S_2 < \cdots < S_n$ and binary labels $R_1, R_2, \ldots, R_n$, we calculate weighted cumulative differences $C_k = \frac{1}{n}\sum_{j=1}^{k}(R_j - S_j)W_j$ for $k=1,2,\ldots,n$:
$$\text{Kuiper} = \max_{0 \leq k \leq n} C_k - \min_{0 \leq k \leq n} C_k$$
where $C_0 = 0$ and $W_j$ are weights. Lower values indicate better calibration. We use $W_j = S_j$ to prioritize calibration in high-confidence regions, reflecting production scenarios where high-confidence judgments are retained while low-confidence cases are delegated to more capable (and costlier) systems or human reviewers.

\paragraph{Baselines.}

We compare to three baselines: verbalized confidence, self-consistency, and majority. \textbf{Verbalized confidence} uses chain-of-thought explanations followed by a verdict and confidence score \citep{xiong2024can, tian2023just}. \textbf{Self-consistency} estimates confidence as the fraction of 
N independent samples voting for a response \citep{wang2022self0consistency}. \textbf{Majority} selects the most frequent verdict among 
N samples, with confidence as the proportion of times the majority answer appears \citep{wang2022self0consistency}. For verbalized confidence, we ablate prompts and score ranges (see \ref{verbalised_ablations}). For consistency and majority, we vary 
$N$ (5--30) and temperature (0--1.5), with optimal results at 
$N=10$, temperature $=0.7$ (see ablations in \ref{temp_calib_perf}--~\ref{num_runs_calib_perf}).

\section{Results}

\begin{figure*}[h]
    \centering

    \begin{subfigure}[t]{\textwidth}
        \centering
        \includegraphics[width=0.7\textwidth]{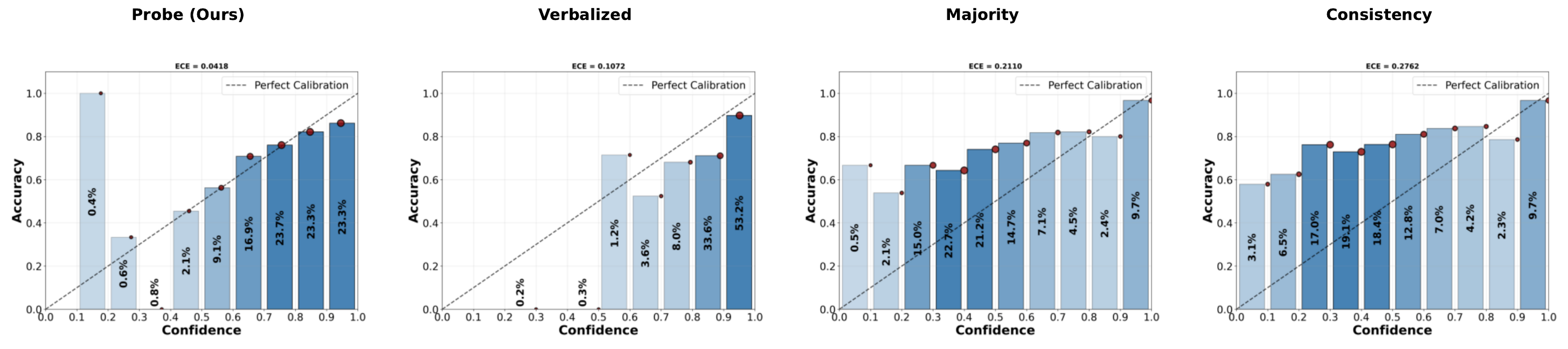}
        \label{fig:ece_plots_a}
    \end{subfigure}


    \begin{subfigure}[t]{\textwidth}
        \centering
        \includegraphics[width=0.7\textwidth]{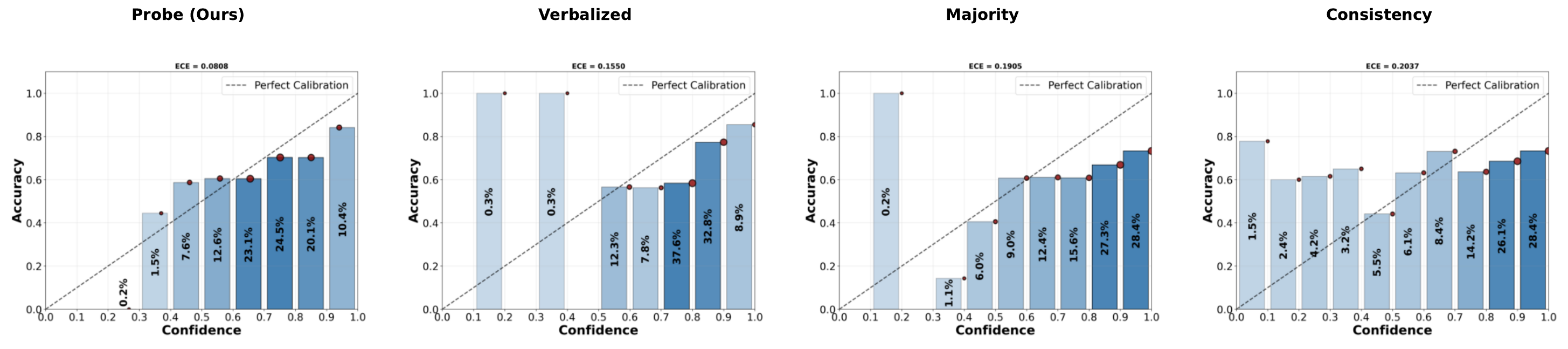}
        \label{fig:ece_plots_b}
    \end{subfigure}

    \caption{Reliability diagrams using 10 bins for Qwen 32B (Prompted Judge) and J1 LLAMA 70B (Finetuned Judge) on JudgeBench. We notice probes generally improve calibration. The color and percentage in each bar present the proportion of data samples in each bin.  The verbalized method is generally overconfident, while multi-generation methods (consistency, majority) may be underconfident (Qwen) or overconfident (LLAMA) depending on model families. 
        Similar trends are observed in finetuned and prompted variants of the judges, as shown in Appendix~\ref{app:reliability_diagrams}.}
    \label{fig:ece_plots}
\end{figure*}

\begin{table}
  \centering
  \footnotesize
  \begin{tabular}{lcc}
    \hline
    \textbf{Method} & \textbf{LLAMA Scout} & \textbf{GPT-OSS 20B} \\
    \hline
    \multicolumn{3}{c}{\textbf{PPE PREFERENCE}} \\
    \hline
    Verbalized & 0.1179 / 0.0892 & 0.2265 / 0.2233 \\
    Consistency & 0.3533 / 0.3287 & 0.2045 / 0.1922 \\
    Majority & 0.3506 / 0.3246 & 0.2056 / 0.1934 \\
    Probe & \textbf{0.0434} / \textbf{0.0582} & \textbf{0.1461} / \textbf{0.1713} \\
    \hline
    \multicolumn{3}{c}{\textbf{PPE CORRECTNESS}} \\
    \hline
    Verbalized & 0.1931 / 0.1597 & 0.1462 / 0.1520 \\
    Consistency & 0.3212 / 0.2978 & 0.0932 / \textbf{0.0980} \\
    Majority & 0.3197 / 0.2940 & 0.0990 / 0.1034 \\
    Probe & \textbf{0.0643} / \textbf{0.0684} & \textbf{0.0689} / 0.1193 \\
    \hline
    \multicolumn{3}{c}{\textbf{JudgeBench}} \\
    \hline
    Verbalized & 0.1388 / 0.1312 & 0.1276 / 0.1267 \\
    Consistency & 0.3011 / 0.2918 & 0.0756 / 0.1097 \\
    Majority & 0.2955 / 0.2842 & 0.0738 / \textbf{0.0997} \\
    Probe & \textbf{0.0563} / \textbf{0.0766} & \textbf{0.0601} / 0.1263 \\
    \hline
    \multicolumn{3}{c}{\textbf{RewardBench}} \\
    \hline
    Verbalized & 0.0742 / \textbf{0.0671} & 0.0198 / \textbf{0.0191} \\
    Consistency & 0.2233 / 0.2037 & \textbf{0.0165} / 0.0248 \\
    Majority & 0.2178 / 0.1996 & 0.0166 / 0.0282 \\
    Probe & \textbf{0.0683} / 0.0879 & 0.1404 / 0.1917 \\
    \hline
  \end{tabular}
  \caption{MoE model evaluation. Probes consistently outperform baselines for Llama Scout (109B) both in- and out-of-distribution. For GPT-OSS (20B), probes excel in-distribution and on JudgeBench but not on RewardBench, exhibiting the same conservative behavior on RewardBench as observed with dense models.}
  \label{tab:moe_performance}
\end{table}

\paragraph{Performance on Dense Models In-Distribution.}
Table~\ref{tab:iid_performance} summarizes calibration results across dense model families and datasets. Our probes consistently achieve the best calibration, with the lowest Kuiper/ECE scores on both PPE Correctness and PPE Preference tasks. Llama-family models show the largest gains, with 70–92\% improvements over multi-generation methods and 64–87\% improvement over verbalized methods, across both prompted and finetuned variants. For Qwen models, probes deliver the best calibration on PPE Correctness and competitive results on PPE Preference, while requiring 10x less compute than multi-generation approaches. Calibration across task categories in the PPE dataset (see~\ref{PPE_subsets}) demonstrates that probes consistently outperform verbalized uncertainty and perform as well as or better than multi-generation methods, even when model accuracy varies across these subsets (i.e., between more difficult and easier tasks). We also see that finetuned and non-finetuned variants achieve similar calibration despite different accuracy levels (see~\ref{app:accuracy}), demonstrating that high accuracy does not guarantee good calibration.

\paragraph{Performance on Out of Distribution Datasets.} Probes generalize effectively to unseen benchmarks (Table~\ref{tab:ood_performance}), consistently outperforming baselines on JudgeBench across all model families, demonstrating strong robustness under domain shift. However, on RewardBench, probe performance lags behind verbalized confidence for several models. This suggests that while probes excel on challenging tasks, their conservative calibration can underperform on easier datasets where models already achieve high accuracy.

\paragraph{Relationship Between Accuracy and Calibration.} We attribute the relative underperformance of probes on RewardBench to their conservative confidence estimation. RewardBench exhibits higher overall accuracy across models, leading verbalized confidence (which tends to overestimate certainty) to perform deceptively well. In contrast, probes are designed to temper overconfidence, producing smoother and more cautious probability distributions.

By analyzing model accuracy at different confidence thresholds in Appendix~\ref{app:accuracy_vs_confidence}, we observe that probes yield higher accuracy among the most confident predictions, but fewer samples are assigned high confidence. Verbalized confidence, by contrast, spreads high confidence too liberally, leading to apparent calibration gains on easy datasets but poor reliability on harder ones.

In safety-critical applications such as medical advice, legal reasoning, or financial decision-making, where false positives are costly, this conservative behavior is highly desirable. Probes could present a favorable trade-off between accuracy and risk, offering more trustworthy confidence estimates even if slightly underconfident on easier datasets.

\paragraph{Verbalized confidence is often the second best option.} Verbalized Confidence is Often the Second Best Option. Among our three baselines, verbalized confidence consistently performs second best after probes. In contrast, multigeneration methods can skew model confidence distributions—making models overconfident (LLaMA) or underconfident (Qwen, GPT-OSS) depending on the model family. We also note that prompting models to reason through uncertainty with hedging phrases yields more calibrated confidence (Appendix~\ref{verbalised_ablations}), supporting the hypothesis that verbally expressing uncertainty improves calibration~\citep{lin2023teaching, lyu2025calibrating}.

\paragraph{Performance on MOE models.}
Table~\ref{tab:moe_performance} extends our evaluation to Mixture-of-Experts (MoE) architectures, LLAMA Scout and GPT-OSS 20B. Probes consistently outperform all baselines across both in-distribution and out-of-distribution datasets, except on RewardBench, where GPT-OSS occasionally shows marginally better Kuiper values for the consistency method. However, probes remain competitive while needing 10× fewer inference calls than consistency, offering an efficiency gain.

\paragraph{Ablations with different losses and layers to train the probe.} We conduct ablation studies in Appendix~\ref{app:focal_loss_ablation} comparing different loss functions for probe training, evaluating binary cross-entropy, focal loss \cite{lin2017focal}, and Brier score loss. We also experiment with probes at various layers (see~\ref{app:layer}). Our results demonstrate that Brier score loss at the middle layer of models yields the most well-calibrated uncertainty estimates.

\paragraph{Single-Pass Selective Classification Baselines.} In Appendix~\ref{selective_classification} we compare against single-pass selective classification baselines—including Perplexity, Maximum Softmax Probability, Energy \cite{liu2020energy0based}, Chain of Embedding \cite{wang2024latent}, CoT-kinetics \cite{bi2025cot0kinetics0}---which use AUROC to measure how well uncertainty scores distinguish correct from incorrect answers. As shown, our probes achieve superior AUROC performance. Additionally, while some baselines require storing hidden states for every token ($\mathcal{O}(\text{tokens} \times \text{layers} \times \text{hidden\_dim})$), our probes only require $\mathcal{O}(\text{layers} \times \text{hidden\_dim})$ memory.

\section{Conclusion}
We present a practical approach to calibrate LLM judges using linear probes on model representations, achieving strong calibration without additional training, multi-sample generation, or per-token storage, enabling plug-and-play, cost-effective deployment. Probe-based calibration significantly outperforms verbalized and multi-generation methods across models, architectures, and tasks, delivering order-of-magnitude computational savings and strong out-of-distribution generalization for industry-scale deployment.

\section*{Limitations} 
Future work could involve more investigation into interpreting which features are learned by the probes and further experimentation to optimize probe generalizability.

One limitation of this approach is that it requires access to the judge's hidden states in the middle layers and a labeled dataset with ground truth verdicts. The method doesn't work for domains where ground truth verdicts aren't available.
Also, in this work, we train one probe for each judge model, but if the judge is retrained or finetuned, the probe may also need to be retrained as the information stored in the hidden states could change.


\bibliography{custom}

@article{zheng2023judging,
	title        = {Judging llm-as-a-judge with mt-bench and chatbot arena},
	author       = {Zheng, Lianmin and Chiang, Wei-Lin and Sheng, Ying and Zhuang, Siyuan and Wu, Zhanghao and Zhuang, Yonghao and Lin, Zi and Li, Zhuohan and Li, Dacheng and Xing, Eric and others},
	year         = 2023,
	journal      = {Advances in neural information processing systems},
	volume       = 36,
	pages        = {46595--46623}
}

@inproceedings{lyu2025calibrating,
	title        = {Calibrating large language models with sample consistency},
	author       = {Lyu, Qing and Shridhar, Kumar and Malaviya, Chaitanya and Zhang, Li and Elazar, Yanai and Tandon, Niket and Apidianaki, Marianna and Sachan, Mrinmaya and Callison-Burch, Chris},
	year         = 2025,
	booktitle    = {Proceedings of the AAAI Conference on Artificial Intelligence},
	volume       = 39,
	number       = 18,
	pages        = {19260--19268}
}

@article{lee2023rlaif,
	title        = {RLAIF vs. RLHF: Scaling Reinforcement Learning from Human Feedback with AI Feedback},
	author       = {Harrison Lee and Samrat Phatale and Hassan Mansoor and Thomas Mesnard and Johan Ferret and Kellie Lu and Colton Bishop and Ethan Hall and Victor Carbune and Abhinav Rastogi and Sushant Prakash},
	year         = 2023,
	journal      = {arXiv preprint arXiv: 2309.00267}
}

@article{ouyang2022training,
	title        = {Training language models to follow instructions with human feedback},
	author       = {Ouyang, Long and Wu, Jeffrey and Jiang, Xu and Almeida, Diogo and Wainwright, Carroll and Mishkin, Pamela and Zhang, Chong and Agarwal, Sandhini and Slama, Katarina and Ray, Alex and others},
	year         = 2022,
	journal      = {Advances in Neural Information Processing Systems},
	volume       = 35,
	pages        = {27730--27744}
}

@article{tan2024judgebench0,
	title        = {JudgeBench: A Benchmark for Evaluating LLM-based Judges},
	author       = {Sijun Tan and Siyuan Zhuang and Kyle Montgomery and William Y. Tang and Alejandro Cuadron and Chenguang Wang and R. Popa and Ion Stoica},
	year         = 2024,
	journal      = {International Conference on Learning Representations},
	doi          = {10.48550/arXiv.2410.12784},
	bibsource    = {Semantic Scholar https://www.semanticscholar.org/paper/088ab579bf490691eea7ac92e122ee11c9b9d131}
}

@article{lambert2024rewardbench0,
	title        = {RewardBench: Evaluating Reward Models for Language Modeling},
	author       = {Nathan Lambert and Valentina Pyatkin and Jacob Morrison and LJ Miranda and Bill Yuchen Lin and Khyathi Chandu and Nouha Dziri and Sachin Kumar and Tom Zick and Yejin Choi and Noah A. Smith and Hannaneh Hajishirzi},
	year         = 2024,
	journal      = {arXiv preprint arXiv: 2403.13787}
}

@article{frick2024evaluate,
	title        = {How to Evaluate Reward Models for RLHF},
	author       = {Evan Frick and Tianle Li and Connor Chen and Wei-Lin Chiang and Anastasios Nikolas Angelopoulos and Jiantao Jiao and Banghua Zhu and Joseph Gonzalez and I. Stoica},
	year         = 2024,
	journal      = {International Conference on Learning Representations},
	doi          = {10.48550/arXiv.2410.14872},
	bibsource    = {Semantic Scholar https://www.semanticscholar.org/paper/f2fde6be4b074f509cf974d1aac24019247473ae}
}

@article{bi2025cot0kinetics0,
	title        = {CoT-Kinetics: A Theoretical Modeling Assessing LRM Reasoning Process},
	author       = {Jinhe Bi and Danqi Yan and Yifan Wang and Wenke Huang and Haokun Chen and Guancheng Wan and Mang Ye and Xun Xiao and Hinrich Schuetze and Volker Tresp and Yunpu Ma},
	year         = 2025,
	journal      = {arXiv preprint arXiv: 2505.13408}
}

@article{wang2024latent,
	title        = {Latent space chain-of-embedding enables output-free llm self-evaluation},
	author       = {Wang, Yiming and Zhang, Pei and Yang, Baosong and Wong, Derek F and Wang, Rui},
	year         = 2024,
	journal      = {arXiv preprint arXiv:2410.13640}
}

@article{ji2025calibrating,
	title        = {Calibrating Verbal Uncertainty as a Linear Feature to Reduce Hallucinations},
	author       = {Ziwei Ji and Lei Yu and Yeskendir Koishekenov and Yejin Bang and Anthony Hartshorn and Alan Schelten and Cheng Zhang and Pascale Fung and Nicola Cancedda},
	year         = 2025,
	journal      = {arXiv preprint arXiv: 2503.14477}
}

@inproceedings{sriramanan2024llmcheck,
	title        = {{LLM}-Check: Investigating Detection of Hallucinations in Large Language Models},
	author       = {Gaurang Sriramanan and Siddhant Bharti and Vinu Sankar Sadasivan and Shoumik Saha and Priyatham Kattakinda and Soheil Feizi},
	year         = 2024,
	booktitle    = {The Thirty-eighth Annual Conference on Neural Information Processing Systems},
	url          = {https://openreview.net/forum?id=LYx4w3CAgy}
}

@inproceedings{xiong2024can,
	title        = {Can {LLM}s Express Their Uncertainty? An Empirical Evaluation of Confidence Elicitation in {LLM}s},
	author       = {Miao Xiong and Zhiyuan Hu and Xinyang Lu and YIFEI LI and Jie Fu and Junxian He and Bryan Hooi},
	year         = 2024,
	booktitle    = {The Twelfth International Conference on Learning Representations},
	url          = {https://openreview.net/forum?id=gjeQKFxFpZ}
}

@article{damani2025beyond,
	title        = {Beyond binary rewards: Training lms to reason about their uncertainty},
	author       = {Damani, Mehul and Puri, Isha and Slocum, Stewart and Shenfeld, Idan and Choshen, Leshem and Kim, Yoon and Andreas, Jacob},
	year         = 2025,
	journal      = {arXiv preprint arXiv:2507.16806}
}

@article{jung2024trust,
	title        = {Trust or escalate: Llm judges with provable guarantees for human agreement},
	author       = {Jung, Jaehun and Brahman, Faeze and Choi, Yejin},
	year         = 2024,
	journal      = {arXiv preprint arXiv:2407.18370}
}

@article{dubois2023alpacafarm,
	title        = {Alpacafarm: A simulation framework for methods that learn from human feedback},
	author       = {Dubois, Yann and Li, Chen Xuechen and Taori, Rohan and Zhang, Tianyi and Gulrajani, Ishaan and Ba, Jimmy and Guestrin, Carlos and Liang, Percy S and Hashimoto, Tatsunori B},
	year         = 2023,
	journal      = {Advances in Neural Information Processing Systems},
	volume       = 36,
	pages        = {30039--30069}
}

@article{bai2022constitutional,
	title        = {Constitutional ai: Harmlessness from ai feedback},
	author       = {Bai, Yuntao and Kadavath, Saurav and Kundu, Sandipan and Askell, Amanda and Kernion, Jackson and Jones, Andy and Chen, Anna and Goldie, Anna and Mirhoseini, Azalia and McKinnon, Cameron and others},
	year         = 2022,
	journal      = {arXiv preprint arXiv:2212.08073}
}

@article{christiano2017deep,
	title        = {Deep reinforcement learning from human preferences},
	author       = {Christiano, Paul F and Leike, Jan and Brown, Tom and Martic, Miljan and Legg, Shane and Amodei, Dario},
	year         = 2017,
	journal      = {Advances in Neural Information Processing Systems},
	volume       = 30
}

@article{kim2024prometheus2,
	title        = {Prometheus 2: An Open Source Language Model Specialized in Evaluating Other Language Models},
	author       = {Kim, Seungone and Shin, Juyoung and Kwak, Yejin and Zhao, Yujin and Jang, Jungwoo and Hwang, Ahra and Kim, Geon and Kim, Hyeonbin and Yoon, Junho and Park, Hahyeon and others},
	year         = 2024,
	journal      = {arXiv preprint arXiv:2405.01535}
}

@article{chen2023frugalgpt,
	title        = {FrugalGPT: How to Use Large Language Models While Reducing Cost and Improving Performance},
	author       = {Chen, Lingjiao and Zaharia, Matei and Zou, James},
	year         = 2023,
	journal      = {arXiv preprint arXiv:2305.05176}
}

@article{zhao2021calibrate,
	title        = {Calibrate Before Use: Improving Few-Shot Performance of Language Models},
	author       = {Zhao, Tony Z and Wallace, Eric and Feng, Shi and Klein, Dan and Singh, Sameer},
	year         = 2021,
	journal      = {International Conference on Machine Learning},
	pages        = {12697--12706}
}

@article{kuhn2023semantic,
	title        = {Semantic Uncertainty: Linguistic Invariances for Uncertainty Estimation in Natural Language Generation},
	author       = {Kuhn, Lorenz and Gal, Yarin and Farquhar, Sebastian},
	year         = 2023,
	journal      = {International Conference on Learning Representations}
}

@article{wang2022self0consistency,
	title        = {Self-Consistency Improves Chain of Thought Reasoning in Language Models},
	author       = {Xuezhi Wang and Jason Wei and Dale Schuurmans and Quoc Le and Ed Chi and Sharan Narang and Aakanksha Chowdhery and Denny Zhou},
	year         = 2022,
	journal      = {arXiv preprint arXiv: 2203.11171}
}

@inproceedings{hongli-etal-2024-mitigating,
	title        = {Mitigating the Bias of Large Language Model Evaluation},
	author       = {Zhou, Hongli  and Huang, Hui  and Long, Yunfei  and Xu, Bing  and Zhu, Conghui  and Cao, Hailong  and Yang, Muyun  and Zhao, Tiejun},
	year         = 2024,
	month        = jul,
	booktitle    = {Proceedings of the 23rd Chinese National Conference on Computational Linguistics (Volume 1: Main Conference)},
	publisher    = {Chinese Information Processing Society of China},
	address      = {Taiyuan, China},
	pages        = {1310--1319},
	url          = {https://aclanthology.org/2024.ccl-1.101/},
	editor       = {Maosong, Sun  and Jiye, Liang  and Xianpei, Han  and Zhiyuan, Liu  and Yulan, He}
}

@inproceedings{nie-etal-2020-learn,
	title        = {What Can We Learn from Collective Human Opinions on Natural Language Inference Data?},
	author       = {Nie, Yixin  and Zhou, Xiang  and Bansal, Mohit},
	year         = 2020,
	month        = nov,
	booktitle    = {Proceedings of the 2020 Conference on Empirical Methods in Natural Language Processing (EMNLP)},
	publisher    = {Association for Computational Linguistics},
	address      = {Online},
	pages        = {9131--9143},
	doi          = {10.18653/v1/2020.emnlp-main.734},
	url          = {https://aclanthology.org/2020.emnlp-main.734/},
	editor       = {Webber, Bonnie  and Cohn, Trevor  and He, Yulan  and Liu, Yang}
}

@article{pavlick-kwiatkowski-2019-inherent,
	title        = {Inherent Disagreements in Human Textual Inferences},
	author       = {Pavlick, Ellie  and Kwiatkowski, Tom},
	year         = 2019,
	journal      = {Transactions of the Association for Computational Linguistics},
	publisher    = {MIT Press},
	address      = {Cambridge, MA},
	volume       = 7,
	pages        = {677--694},
	doi          = {10.1162/tacl_a_00293},
	url          = {https://aclanthology.org/Q19-1043/},
	editor       = {Lee, Lillian  and Johnson, Mark  and Roark, Brian  and Nenkova, Ani}
}

@inproceedings{ross2024what,
	title        = {What makes a good metric? Evaluating automatic metrics for text-to-image consistency},
	author       = {Candace Ross and Melissa Hall and Adriana Romero-Soriano and Adina Williams},
	year         = 2024,
	booktitle    = {First Conference on Language Modeling},
	url          = {https://openreview.net/forum?id=LFfktMPAci}
}

@inproceedings{chen-mueller-2024-quantifying,
	title        = {Quantifying Uncertainty in Answers from any Language Model and Enhancing their Trustworthiness},
	author       = {Chen, Jiuhai and Mueller, Jonas},
	year         = 2024,
	month        = {aug},
	booktitle    = {Proceedings of the 62nd Annual Meeting of the Association for Computational Linguistics (Volume 1: Long Papers)},
	publisher    = {Association for Computational Linguistics},
	address      = {Bangkok, Thailand},
	pages        = {5186--5200},
	doi          = {10.18653/v1/2024.acl-long.283},
	url          = {https://aclanthology.org/2024.acl-long.283/},
	editor       = {Ku, Lun-Wei and Martins, Andre and Srikumar, Vivek}
}

@inproceedings{kapoor-etal-2024-calibration,
	title        = {Calibration-Tuning: Teaching Large Language Models to Know What They Don{'}t Know},
	author       = {Kapoor, Sanyam  and Gruver, Nate  and Roberts, Manley  and Pal, Arka  and Dooley, Samuel  and Goldblum, Micah  and Wilson, Andrew},
	year         = 2024,
	month        = mar,
	booktitle    = {Proceedings of the 1st Workshop on Uncertainty-Aware NLP (UncertaiNLP 2024)},
	publisher    = {Association for Computational Linguistics},
	address      = {St Julians, Malta},
	pages        = {1--14},
	url          = {https://aclanthology.org/2024.uncertainlp-1.1/},
	editor       = {V{\'a}zquez, Ra{\'u}l  and Celikkanat, Hande  and Ulmer, Dennis  and Tiedemann, J{\"o}rg  and Swayamdipta, Swabha  and Aziz, Wilker  and Plank, Barbara  and Baan, Joris  and de Marneffe, Marie-Catherine}
}

@article{burns2023discovering,
	title        = {Discovering Latent Knowledge in Language Models Without Supervision},
	author       = {Burns, Collin and Ye, Haotian and Klein, Dan and Steinhardt, Jacob},
	year         = 2023,
	journal      = {International Conference on Learning Representations}
}

@article{zhang2025reasoning,
	title        = {Reasoning Models Know When They're Right: Probing Hidden States for Self-Verification},
	author       = {Anqi Zhang and Yulin Chen and Jane Pan and Chen Zhao and Aurojit Panda and Jinyang Li and He He},
	year         = 2025,
	journal      = {arXiv preprint arXiv: 2504.05419}
}

@article{azaria2023internal,
	title        = {The Internal State of an LLM Knows When It's Lying},
	author       = {Azaria, Amos and Mitchell, Tom M},
	year         = 2023,
	journal      = {Findings of the Association for Computational Linguistics}
}

@article{kadavath2022language,
	title        = {Language Models (Mostly) Know What They Know},
	author       = {Saurav Kadavath and Tom Conerly and Amanda Askell and Tom Henighan and Dawn Drain and Ethan Perez and Nicholas Schiefer and Zac Hatfield-Dodds and Nova DasSarma and Eli Tran-Johnson and Scott Johnston and Sheer El-Showk and Andy Jones and Nelson Elhage and Tristan Hume and Anna Chen and Yuntao Bai and Sam Bowman and Stanislav Fort and Deep Ganguli and Danny Hernandez and Josh Jacobson and Jackson Kernion and Shauna Kravec and Liane Lovitt and Kamal Ndousse and Catherine Olsson and Sam Ringer and Dario Amodei and Tom Brown and Jack Clark and Nicholas Joseph and Ben Mann and Sam McCandlish and Chris Olah and Jared Kaplan},
	year         = 2022,
	journal      = {arXiv preprint arXiv: 2207.05221}
}

@article{radharapu2025arbiters,
	title        = {Arbiters of Ambivalence: Challenges of Using LLMs in No-Consensus Tasks},
	author       = {Bhaktipriya Radharapu and Manon Revel and Megan Ung and Sebastian Ruder and Adina Williams},
	year         = 2025,
	journal      = {arXiv preprint arXiv: 2505.23820}
}

@article{Aroyo_Welty_2015,
	title        = {Truth Is a Lie: Crowd Truth and the Seven Myths of Human Annotation},
	author       = {Aroyo, Lora and Welty, Chris},
	year         = 2015,
	month        = {Mar.},
	journal      = {AI Magazine},
	volume       = 36,
	number       = 1,
	pages        = {15--24},
	doi          = {10.1609/aimag.v36i1.2564},
	url          = {https://ojs.aaai.org/aimagazine/index.php/aimagazine/article/view/2564},
	abstractnote = {Big data is having a disruptive impact across the sciences. Human annotation of semantic interpretation tasks is a critical part of big data semantics, but it is based on an antiquated ideal of a single correct truth that needs to be similarly disrupted. We expose seven myths about human annotation, most of which derive from that antiquated ideal of truth, and dispell these myths with examples from our research. We propose a new theory of truth, crowd truth, that is based on the intuition that human interpretation is subjective, and that measuring annotations on the same objects of interpretation (in our examples, sentences) across a crowd will provide a useful representation of their subjectivity and the range of reasonable interpretations.}
}

@inproceedings{chiang2024chatbot,
	title        = {Chatbot arena: An open platform for evaluating llms by human preference},
	author       = {Chiang, Wei-Lin and Zheng, Lianmin and Sheng, Ying and Angelopoulos, Anastasios Nikolas and Li, Tianle and Li, Dacheng and Zhu, Banghua and Zhang, Hao and Jordan, Michael and Gonzalez, Joseph E and others},
	year         = 2024,
	booktitle    = {Forty-first International Conference on Machine Learning}
}

@article{kim2024prometheus,
	title        = {Prometheus: Inducing fine-grained evaluation capability in language models},
	author       = {Kim, Seungone and Shin, Juyoung and Cho, Yejin and Jang, Joel and Longpre, Shayne and Lee, Hwaran and Yun, Sangdoo and Shin, Sungdong and Kim, Seongjin and Thorne, James and others},
	year         = 2024,
	journal      = {arXiv preprint arXiv:2310.08491}
}

@misc{zhu2023judgelm,
	title        = {JudgeLM: Fine-tuned Large Language Models are Scalable Judges},
	author       = {Lianghui Zhu and Xinggang Wang and Xinlong Wang},
	year         = 2025,
	url          = {https://arxiv.org/abs/2310.17631},
	eprint       = {2310.17631},
	archiveprefix = {arXiv},
	primaryclass = {cs.CL}
}

@article{li2024generative,
	title        = {Generative judge for evaluating alignment},
	author       = {Li, Junlong and Sun, Shichao and Yuan, Weizhe and Fan, Run-Ze and Zhao, Hai and Liu, Pengfei},
	year         = 2024,
	journal      = {arXiv preprint arXiv:2310.05470}
}

@article{wang2023large,
	title        = {Large language models are not fair evaluators},
	author       = {Wang, Peiyi and Li, Lei and Chen, Liang and Zhu, Dawei and Lin, Binghuai and Cao, Yunbo and Liu, Qi and Liu, Tianyu and Sui, Zhifang},
	year         = 2023,
	journal      = {arXiv preprint arXiv:2305.17926}
}

@article{tian2025overconfidence,
	title        = {Overconfidence in LLM-as-a-Judge: Diagnosis and Confidence-Driven Solution},
	author       = {Zailong Tian and Zhuoheng Han and Yanzhe Chen and Haozhe Xu and Xi Yang and Richeng Xuan and Houfeng Wang and Lizi Liao},
	year         = 2025,
	journal      = {arXiv preprint arXiv: 2508.06225}
}

@article{gao2023scaling,
	title        = {Scaling Laws for Reward Model Overoptimization},
	author       = {Leo Gao and John Schulman and Jacob Hilton},
	year         = 2022,
	journal      = {arXiv preprint arXiv: 2210.10760}
}

@article{kirk2024prism,
	title        = {The PRISM alignment dataset: What participatory, representative and individualised human feedback reveals about the subjective and multicultural alignment of large language models},
	author       = {Kirk, Hannah Rose and Whitefield, Alexander and Rottger, Paul and Bean, Andrew M and Margatina, Katerina and Mosquera-Gomez, Rafael and Ciro, Juan and Bartolo, Max and Williams, Adina and He, He and others},
	year         = 2024,
	journal      = {Advances in Neural Information Processing Systems},
	volume       = 37,
	pages        = {105236--105344}
}

@article{openai2024o1,
	title        = {Learning to reason with llms},
	author       = {{OpenAI}},
	year         = 2024,
	journal      = {OpenAI Blog},
	url          = {https://openai.com/index/learning-to-reason-with-llms/}
}

@article{deepseek2025r1,
	title        = {Deepseek-r1: Incentivizing reasoning capability in llms via reinforcement learning},
	author       = {{DeepSeek-AI}},
	year         = 2025,
	journal      = {arXiv preprint arXiv:2501.12948}
}

@inproceedings{tian2023just,
	title        = {Just Ask for Calibration: Strategies for Eliciting Calibrated Confidence Scores from Language Models Fine-Tuned with Human Feedback},
	author       = {Tian, Katherine and Mitchell, Eric and Zhou, Allan and Sharma, Archit and Rafailov, Rafael and Yao, Huaxiu and Finn, Chelsea and Manning, Christopher},
	year         = 2023,
	month        = {dec},
	booktitle    = {Proceedings of the 2023 Conference on Empirical Methods in Natural Language Processing},
	publisher    = {Association for Computational Linguistics},
	address      = {Singapore},
	pages        = {5433--5442},
	doi          = {10.18653/v1/2023.emnlp-main.330},
	url          = {https://aclanthology.org/2023.emnlp-main.330/},
	editor       = {Bouamor, Houda and Pino, Juan and Bali, Kalika}
}

@article{lin2023teaching,
	title        = {Teaching Models to Express Their Uncertainty in Words},
	author       = {Stephanie Lin and Jacob Hilton and Owain Evans},
	year         = 2022,
	journal      = {arXiv preprint arXiv: 2205.14334}
}

@article{guo2017calibration,
	title        = {On Calibration of Modern Neural Networks},
	author       = {Chuan Guo and Geoff Pleiss and Yu Sun and Kilian Q. Weinberger},
	year         = 2017,
	journal      = {arXiv preprint arXiv: 1706.04599}
}

@inproceedings{zhou2024batch,
	title        = {Batch Calibration: Rethinking Calibration for In-Context Learning and Prompt Engineering},
	author       = {Han Zhou and Xingchen Wan and Lev Proleev and Diana Mincu and Jilin Chen and Katherine A Heller and Subhrajit Roy},
	year         = 2024,
	booktitle    = {The Twelfth International Conference on Learning Representations},
	url          = {https://openreview.net/forum?id=L3FHMoKZcS}
}

@article{li2025conftuner0,
	title        = {ConfTuner: Training Large Language Models to Express Their Confidence Verbally},
	author       = {Yibo Li and Miao Xiong and Jiaying Wu and Bryan Hooi},
	year         = 2025,
	journal      = {arXiv preprint arXiv: 2508.18847}
}

@article{huang2023look,
	title        = {Look before you leap: An exploratory study of uncertainty measurement for large language models},
	author       = {Huang, Yuheng and Song, Jiayang and Wang, Zhijie and Chen, Huaming and Shen, Likun and Lei, Xin and Wang, Felix Juefei-Xu},
	year         = 2023,
	journal      = {arXiv preprint arXiv:2307.10236}
}

@article{zou2023representation,
	title        = {Representation engineering: A top-down approach to ai transparency},
	author       = {Zou, Andy and Phan, Long and Chen, Sarah and Campbell, James and Guo, Phillip and Ren, Richard and Pan, Alexander and Yin, Xuwang and Mazeika, Mantas and Dombrowski, Ann-Kathrin and others},
	year         = 2023,
	journal      = {arXiv preprint arXiv:2310.01405}
}

@article{li2024inference,
	title        = {Inference-time intervention: Eliciting truthful answers from a language model},
	author       = {Li, Kenneth and Patel, Oam and Vi{\'e}gas, Fernanda and Pfister, Hanspeter and Wattenberg, Martin},
	year         = 2024,
	journal      = {Advances in Neural Information Processing Systems},
	volume       = 36
}

@article{manakul2023selfcheckgpt,
	title        = {Selfcheckgpt: Zero-resource black-box hallucination detection for generative large language models},
	author       = {Manakul, Potsawee and Liusie, Adian and Gales, Mark JF},
	year         = 2023,
	journal      = {arXiv preprint arXiv:2303.08896}
}

@article{kossen2024semantic,
	title        = {Semantic Entropy Probes: Robust and Cheap Hallucination Detection in LLMs},
	author       = {Jannik Kossen and Jiatong Han and Muhammed Razzak and Lisa Schut and Shreshth Malik and Yarin Gal},
	year         = 2024,
	journal      = {arXiv preprint arXiv: 2406.15927}
}

@misc{touvron2023llama2openfoundation,
	title        = {Llama 2: Open Foundation and Fine-Tuned Chat Models},
	author       = {Hugo Touvron and Louis Martin and Kevin Stone and Peter Albert and Amjad Almahairi and Yasmine Babaei and Nikolay Bashlykov and Soumya Batra and Prajjwal Bhargava and Shruti Bhosale and Dan Bikel and Lukas Blecher and Cristian Canton Ferrer and Moya Chen and Guillem Cucurull and David Esiobu and Jude Fernandes and Jeremy Fu and Wenyin Fu and Brian Fuller and Cynthia Gao and Vedanuj Goswami and Naman Goyal and Anthony Hartshorn and Saghar Hosseini and Rui Hou and Hakan Inan and Marcin Kardas and Viktor Kerkez and Madian Khabsa and Isabel Kloumann and Artem Korenev and Punit Singh Koura and Marie-Anne Lachaux and Thibaut Lavril and Jenya Lee and Diana Liskovich and Yinghai Lu and Yuning Mao and Xavier Martinet and Todor Mihaylov and Pushkar Mishra and Igor Molybog and Yixin Nie and Andrew Poulton and Jeremy Reizenstein and Rashi Rungta and Kalyan Saladi and Alan Schelten and Ruan Silva and Eric Michael Smith and Ranjan Subramanian and Xiaoqing Ellen Tan and Binh Tang and Ross Taylor and Adina Williams and Jian Xiang Kuan and Puxin Xu and Zheng Yan and Iliyan Zarov and Yuchen Zhang and Angela Fan and Melanie Kambadur and Sharan Narang and Aurelien Rodriguez and Robert Stojnic and Sergey Edunov and Thomas Scialom},
	year         = 2023,
	url          = {https://arxiv.org/abs/2307.09288},
	eprint       = {2307.09288},
	archiveprefix = {arXiv},
	primaryclass = {cs.CL}
}

@inproceedings{talat-etal-2022-machine,
	title        = {On the Machine Learning of Ethical Judgments from Natural Language},
	author       = {Talat, Zeerak  and Blix, Hagen  and Valvoda, Josef  and Ganesh, Maya Indira  and Cotterell, Ryan  and Williams, Adina},
	year         = 2022,
	month        = jul,
	booktitle    = {Proceedings of the 2022 Conference of the North American Chapter of the Association for Computational Linguistics: Human Language Technologies},
	publisher    = {Association for Computational Linguistics},
	address      = {Seattle, United States},
	pages        = {769--779},
	doi          = {10.18653/v1/2022.naacl-main.56},
	url          = {https://aclanthology.org/2022.naacl-main.56/},
	editor       = {Carpuat, Marine  and de Marneffe, Marie-Catherine  and Meza Ruiz, Ivan Vladimir}
}

@misc{tygert_2025_15253140,
	title        = {Calibration and bias in algorithms, data, and models: a tutorial on metrics and plots for measuring calibration, bias, fairness, reliability, and robustness},
	author       = {Tygert, Mark},
	year         = 2025,
	month        = apr,
	publisher    = {Zenodo},
	doi          = {10.5281/zenodo.15253140},
	url          = {https://doi.org/10.5281/zenodo.15253140}
}

@article{liu2020energy0based,
	title        = {Energy-based Out-of-distribution Detection},
	author       = {Weitang Liu and Xiaoyun Wang and John D. Owens and Yixuan Li},
	year         = 2020,
	journal      = {arXiv preprint arXiv: 2010.03759}
}

@article{whitehouse2025j10,
	title        = {J1: Incentivizing Thinking in LLM-as-a-Judge via Reinforcement Learning},
	author       = {Chenxi Whitehouse and Tianlu Wang and Ping Yu and Xian Li and Jason Weston and Ilia Kulikov and Swarnadeep Saha},
	year         = 2025,
	journal      = {arXiv preprint arXiv: 2505.10320}
}

@misc{qwen3technicalreport,
	title        = {Qwen3 Technical Report},
	author       = {{Qwen Team}},
	year         = 2025,
	url          = {https://arxiv.org/abs/2505.09388},
	eprint       = {2505.09388},
	archiveprefix = {arXiv},
	primaryclass = {cs.CL}
}

@article{llama3modelcard,
	title        = {Llama 3 Model Card},
	author       = {AI@Meta},
	year         = 2024,
	url          = {https://github.com/meta-llama/llama3/blob/main/MODEL_CARD.md}
}

@misc{meta_llama4_scout_17b_16e_instruct_2025,
	title        = {Llama-4-Scout-17B-16E-Instruct},
	author       = {Meta AI},
	year         = 2025,
	note         = {Llama 4 Scout is a 17B parameter, 16 expert mixture-of-experts multimodal model. Licensed under the Llama 4 Community License Agreement.},
	howpublished = {\url{https://huggingface.co/meta-llama/Llama-4-Scout-17B-16E-Instruct}},
	organization = {Meta Platforms, Inc.},
	version      = {Llama 4 Scout, released April 5, 2025}
}

@misc{openai2025gptoss120bgptoss20bmodel,
	title        = {gpt-oss-120b \& gpt-oss-20b Model Card},
	author       = {OpenAI},
	year         = 2025,
	url          = {https://arxiv.org/abs/2508.10925},
	eprint       = {2508.10925},
	archiveprefix = {arXiv},
	primaryclass = {cs.CL}
}

@article{roelofs2020mitigating,
	title        = {Mitigating Bias in Calibration Error Estimation},
	author       = {Rebecca Roelofs and Nicholas Cain and Jonathon Shlens and Michael C. Mozer},
	year         = 2020,
	journal      = {arXiv preprint arXiv: 2012.08668}
}

@article{nixon2019measuring,
	title        = {Measuring Calibration in Deep Learning},
	author       = {Jeremy Nixon and Mike Dusenberry and Ghassen Jerfel and Timothy Nguyen and Jeremiah Liu and Linchuan Zhang and Dustin Tran},
	year         = 2019,
	journal      = {arXiv preprint arXiv: 1904.01685}
}

@article{li2025judging,
	title        = {Judging with Confidence: Calibrating Autoraters to Preference Distributions},
	author       = {Zhuohang Li and Xiaowei Li and Chengyu Huang and Guowang Li and Katayoon Goshvadi and Bo Dai and Dale Schuurmans and Paul Zhou and Hamid Palangi and Yiwen Song and Palash Goyal and Murat Kantarcioglu and Bradley A. Malin and Yuan Xue},
	year         = 2025,
	journal      = {arXiv preprint arXiv: 2510.00263}
}

@article{tao2025revisiting,
	title        = {Revisiting Uncertainty Estimation and Calibration of Large Language Models},
	author       = {Linwei Tao and Yi-Fan Yeh and Minjing Dong and Tao Huang and Philip Torr and Chang Xu},
	year         = 2025,
	journal      = {arXiv preprint arXiv: 2505.23854}
}

@misc{chen2025judgelrmlargereasoningmodels,
	title        = {JudgeLRM: Large Reasoning Models as a Judge},
	author       = {Nuo Chen and Zhiyuan Hu and Qingyun Zou and Jiaying Wu and Qian Wang and Bryan Hooi and Bingsheng He},
	year         = 2025,
	url          = {https://arxiv.org/abs/2504.00050},
	eprint       = {2504.00050},
	archiveprefix = {arXiv},
	primaryclass = {cs.CL}
}

@article{zhang2024diverging,
	title        = {Diverging Preferences: When do Annotators Disagree and do Models Know?},
	author       = {Michael JQ Zhang and Zhilin Wang and Jena D. Hwang and Yi Dong and Olivier Delalleau and Yejin Choi and Eunsol Choi and Xiang Ren and Valentina Pyatkin},
	year         = 2024,
	journal      = {arXiv preprint arXiv: 2410.14632}
}

@article{plaut2024probabilities,
	title        = {Probabilities of Chat LLMs Are Miscalibrated but Still Predict Correctness on Multiple-Choice Q\&A},
	author       = {Benjamin Plaut and Nguyen X. Khanh and Tu Trinh},
	year         = 2024,
	journal      = {arXiv preprint arXiv: 2402.13213}
}

@article{xie2024calibrating,
	title        = {Calibrating Language Models with Adaptive Temperature Scaling},
	author       = {Johnathan Xie and Annie S. Chen and Yoonho Lee and Eric Mitchell and Chelsea Finn},
	year         = 2024,
	journal      = {arXiv preprint arXiv: 2409.19817}
}

@inproceedings{NEURIPS2024_d037fd02,
	title        = {Calibrating Reasoning in Language Models with Internal Consistency},
	author       = {Xie, Zhihui and Guo, Jizhou and Yu, Tong and Li, Shuai},
	year         = 2024,
	booktitle    = {Advances in Neural Information Processing Systems},
	publisher    = {Curran Associates, Inc.},
	volume       = 37,
	pages        = {114872--114901},
	url          = {https://proceedings.neurips.cc/paper_files/paper/2024/file/d037fd021c9aace128b8ce25001cdb6c-Paper-Conference.pdf},
	editor       = {A. Globerson and L. Mackey and D. Belgrave and A. Fan and U. Paquet and J. Tomczak and C. Zhang}
}

@article{glenn1950verification,
	title        = {Verification of forecasts expressed in terms of probability},
	author       = {Glenn, W. Brier},
	year         = 1950,
	journal      = {Monthly weather review},
	publisher    = {War Department, Office of the Chief Signal Officer},
	volume       = 78,
	number       = 1,
	pages        = {1--3}
}

@article{fedus2021switch,
	title        = {Switch Transformers: Scaling to Trillion Parameter Models with Simple and Efficient Sparsity},
	author       = {William Fedus and Barret Zoph and Noam Shazeer},
	year         = 2021,
	journal      = {arXiv preprint arXiv: 2101.03961}
}

@inproceedings{lin2017focal,
	title        = {Focal loss for dense object detection},
	author       = {Lin, Tsung-Yi and Goyal, Priya and Girshick, Ross and He, Kaiming and Doll{\'a}r, Piotr},
	year         = 2017,
	booktitle    = {Proceedings of the IEEE international conference on computer vision},
	pages        = {2980--2988}
}

@article{li2024crowdsourced,
	title        = {From Crowdsourced Data to High-Quality Benchmarks: Arena-Hard and BenchBuilder Pipeline},
	author       = {Tianle Li and Wei-Lin Chiang and Evan Frick and Lisa Dunlap and Tianhao Wu and Banghua Zhu and Joseph E. Gonzalez and Ion Stoica},
	year         = 2024,
	journal      = {arXiv preprint arXiv: 2406.11939}
}

@inproceedings{naeini2015obtaining,
	title        = {Obtaining well calibrated probabilities using bayesian binning},
	author       = {Naeini, Mohammad Pezeshk and Cooper, Gregory and Hauskrecht, Milos},
	year         = 2015,
	booktitle    = {Proceedings of the AAAI conference on artificial intelligence (AAAI)},
	volume       = 29,
	number       = 1,
	pages        = {2901--2907}
}

@article{arrieta2022metrics,
	title        = {Metrics of calibration for probabilistic predictions},
	author       = {Arrieta-Ibarra, Imanol and Gujral, Paman and Tannen, Jonathan and Tygert, Mark and Xu, Cherie},
	year         = 2022,
	journal      = {Journal of Machine Learning Research},
	volume       = 23,
	number       = 351,
	pages        = {1--54}
}

\appendix

\onecolumn
\newpage
\section{Appendix}
\label{sec:appendix}

\subsection{Comparison with Other Single Pass Methods}
\label{selective_classification}
\textbf{Selective classification} evaluates how effectively uncertainty scores distinguish between correct and incorrect predictions. The most widely used metric for this is AUROC, which quantifies the likelihood that a correct prediction will have a lower uncertainty score than an incorrect one. An AUROC of 0.5 indicates no discriminative ability (equivalent to random guessing), whereas values approaching 1.0 reflect strong discriminative performance.

Note that some of these methods like \cite{bi2025cot0kinetics0, wang2024latent} outputs are not constrained to the [0, 1] range; instead, they yield arbitrary scores that serve as relative signals—higher values may indicate greater correctness, but these scores are not interpretable as probabilities.

We evaluate various single-pass uncertainty estimation methods on the JudgeBench dataset. These methods provide signals for detecting answer correctness, which we measure using AUROC scores.

Table~\ref{tab:single_pass_baselines} shows the performance comparison across different model architectures.

\begin{table*}[h]
  \centering
  \footnotesize
  \begin{tabularx}{\textwidth}{lXXXXXXX}
    \hline
    \textbf{Method} & \textbf{LLaMA} & \textbf{LLaMA} & \textbf{J1 LLaMA} & \textbf{J1 LLaMA} & \textbf{J1 QWEN} & \textbf{GPT-OSS} & \textbf{QWEN} \\
    \textbf{} & \textbf{8B} & \textbf{70B} & \textbf{8B} & \textbf{70B} & \textbf{32B} & \textbf{OSS 20B} & \textbf{32B} \\
    
    \hline
    MaxProb & 46.77 & 51.48 & 50.67 & 59.61 & 64.27 & 43.54 & 51.47 \\
    Perplexity & 46.44 & 51.31 & 50.81 & 59.62 & 63.88 & 43.15 & 51.20 \\
    Entropy & 46.81 & 51.08 & 50.85 & 59.60 & 64.33 & 46.82 & 50.79 \\
    TempScl & 46.54 & 51.55 & 50.72 & 59.60 & 63.74 & 42.11 & 51.53 \\
    Energy & 47.99 & 50.07 & 53.34 & 50.00 & 49.34 & \textbf{71.53} & 67.79 \\
    CoER & 52.06 & 47.81 & 49.98 & 49.82 & 60.36 & 69.12 & 46.25 \\
    CoEC & 49.99 & 52.24 & 50.56 & 59.13 & 60.26 & 68.46 & 45.99 \\
    CoTK & 46.69 & 51.11 & 50.87 & 59.98 & 62.90 & 63.72 & 48.80 \\
    \hline
    Verbalised & 46.83 & 57.95 & 53.50 & 62.04 & 68.86 & 64.47 & 69.20 \\
    Sem. Entropy & \textbf{54.71} & 58.13 & 54.19 & 60.63 & 66.87 & 64.42 & 65.63 \\
    Probe & 52.38 & \textbf{58.14} & \textbf{55.36} & \textbf{63.17} & \textbf{69.05} & 62.03 & \textbf{71.17} \\
    \hline
  \end{tabularx}
  \caption{AUROC performance on JudgeBench, comparison of single-pass uncertainty estimation methods across model architectures. Higher values indicate better performance. Bold values indicate best performance per model.}
  \label{tab:single_pass_baselines}
\end{table*}

\subsection{Accuracy at Different Thresholds}
\label{app:accuracy_vs_confidence}
\begin{figure*}[h]
    \centering
    \includegraphics[width=\textwidth]{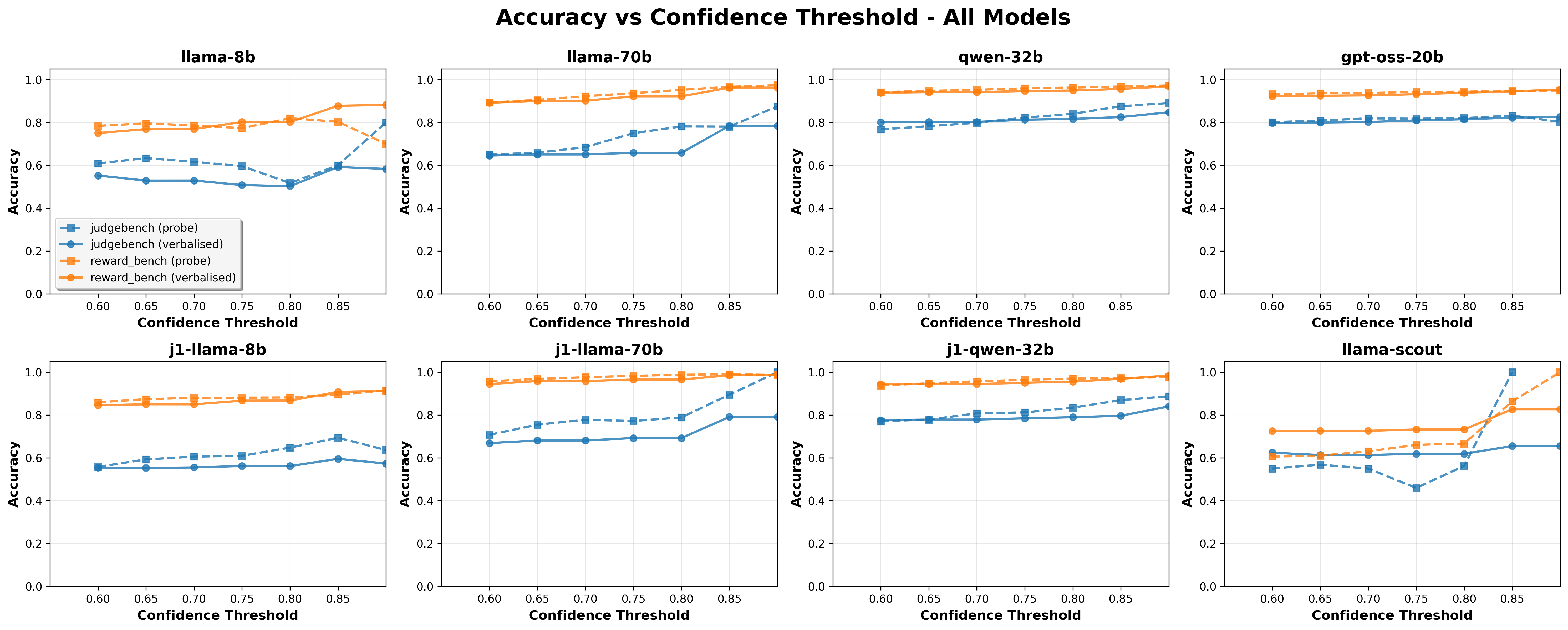}
    \caption{Accuracy of models at different confidence thresholds for Probe and Verbalized uncertainity estimation methods.}
    \label{fig:ece_plots}
\end{figure*}

\subsection{Additional Results --Performance on various PPE Correctness Benchmarks} 
In Table~\ref{tab:dense_ppe_subsets} and Table ~\ref{tab:moe_ppe_subsets}, we present the results on the different subsets of PPE correctness dataset to see if the performance varies by domain. We see that the probe performs well across the board.
\label{PPE_subsets}

\begin{table*}[h]
  \centering
  \footnotesize
    \begin{tabular*}{\textwidth}{@{\extracolsep{\fill}}l*{6}{c}@{}}
    \hline
    \textbf{Method} & \textbf{LLAMA 8B} & \textbf{LLAMA 70B} & \textbf{QWEN 32B} & \textbf{J1 QWEN 32B} & \textbf{J1 LLAMA 8B} & \textbf{J1 LLAMA 70B} \\
    \hline
    \multicolumn{7}{c}{\textbf{PPE Correctness Gpqa}} \\
    \hline
    Verbalized & 0.1846 / 0.1694 & 0.1431 / 0.1365 & 0.1411 / 0.1369 & 0.1444 / 0.1409 & 0.2604 / 0.2508 & 0.1109 / 0.1049 \\
    Consistency & 0.1824 / 0.1857 & 0.2774 / 0.2653 & 0.1136 / 0.1328 & 0.0635 / 0.1075 & 0.3078 / 0.2872 & 0.2270 / 0.2323 \\
    Majority & 0.1792 / 0.1734 & 0.2790 / 0.2623 & 0.1079 / 0.1245 & \textbf{0.0490} / 0.0809 & 0.3086 / 0.2851 & 0.2325 / 0.2175 \\
    Probe & \textbf{0.0457} / \textbf{0.0469} & \textbf{0.0746} / \textbf{0.0717} & \textbf{0.0682} / \textbf{0.0673} & 0.0581 / \textbf{0.0599} & \textbf{0.0782} / \textbf{0.0767} & \textbf{0.0404} / \textbf{0.0444} \\
    \hline
    \multicolumn{7}{c}{\textbf{PPE Correctness Ifeval}} \\
    \hline
    Verbalized & 0.1738 / 0.1690 & 0.1570 / 0.1440 & 0.1738 / 0.1729 & 0.1615 / 0.1603 & 0.2280 / 0.2191 & 0.0788 / 0.0732 \\
    Consistency & 0.1781 / 0.1865 & 0.2244 / 0.2235 & 0.0735 / 0.0985 & 0.0749 / 0.1201 & 0.3729 / 0.3603 & 0.1920 / 0.1836 \\
    Majority & 0.1810 / 0.1704 & 0.2301 / 0.2180 & 0.0708 / 0.0904 & 0.0626 / 0.0986 & 0.3771 / 0.3583 & 0.1946 / 0.1796 \\
    Probe & \textbf{0.0368} / \textbf{0.0437} & \textbf{0.0161} / \textbf{0.0237} & \textbf{0.0596} / \textbf{0.0652} & \textbf{0.0541} / \textbf{0.0586} & \textbf{0.0239} / \textbf{0.0343} & \textbf{0.0393} / \textbf{0.0393} \\
    \hline
    \multicolumn{7}{c}{\textbf{PPE Correctness Math}} \\
    \hline
    Verbalized & 0.1456 / 0.1447 & 0.1349 / 0.1374 & 0.0336 / 0.0362 & 0.0355 / 0.0382 & 0.1259 / 0.1241 & 0.0345 / \textbf{0.0341} \\
    Consistency & 0.0501 / 0.0939 & 0.1025 / 0.1223 & 0.2789 / 0.2950 & 0.2195 / 0.2424 & 0.0843 / 0.0820 & 0.0523 / 0.0905 \\
    Majority & 0.0509 / 0.0725 & 0.1098 / 0.1041 & 0.2747 / 0.2903 & 0.2153 / 0.2365 & 0.0841 / 0.0796 & 0.0384 / 0.0662 \\
    Probe & \textbf{0.0443} / \textbf{0.0459} & \textbf{0.0340} / \textbf{0.0383} & \textbf{0.0336} / \textbf{0.0316} & \textbf{0.0353} / \textbf{0.0349} & \textbf{0.0155} / \textbf{0.0165} & \textbf{0.0310} / \textbf{0.0342} \\
    \hline
    \multicolumn{7}{c}{\textbf{PPE Correctness Mbpp}} \\
    \hline
    Verbalized & 0.2506 / 0.2415 & 0.1905 / 0.1829 & 0.2280 / 0.2333 & 0.2196 / 0.2265 & 0.3073 / 0.2995 & 0.1331 / 0.1291 \\
    Consistency & 0.2573 / 0.2364 & 0.2995 / 0.2762 & \textbf{0.0255} / 0.0412 & 0.0690 / 0.0976 & 0.3943 / 0.3782 & 0.1538 / 0.1606 \\
    Majority & 0.2511 / 0.2276 & 0.2929 / 0.2743 & 0.0293 / \textbf{0.0365} & 0.0747 / 0.0751 & 0.3947 / 0.3829 & 0.1578 / 0.1452 \\
    Probe & \textbf{0.0486} / \textbf{0.0471} & \textbf{0.0445} / \textbf{0.0644} & 0.0366 / 0.0379 & \textbf{0.0215} / \textbf{0.0401} & \textbf{0.0335} / \textbf{0.0406} & \textbf{0.0322} / \textbf{0.0551} \\
    \hline
    \multicolumn{7}{c}{\textbf{PPE Correctness Mmlu}} \\
    \hline
    Verbalized & 0.1546 / 0.1469 & 0.0607 / 0.0608 & 0.0470 / 0.0475 & 0.0357 / 0.0405 & 0.1779 / 0.1766 & 0.0430 / 0.0422 \\
    Consistency & 0.1209 / 0.1533 & 0.1586 / 0.1697 & 0.1869 / 0.2000 & 0.1439 / 0.1662 & 0.2157 / 0.2032 & 0.1046 / 0.1156 \\
    Majority & 0.1235 / 0.1315 & 0.1674 / 0.1603 & 0.1843 / 0.1994 & 0.1329 / 0.1506 & 0.2164 / 0.2063 & 0.1077 / 0.1015 \\
    Probe & \textbf{0.0386} / \textbf{0.0406} & \textbf{0.0287} / \textbf{0.0304} & \textbf{0.0202} / \textbf{0.0277} & \textbf{0.0291} / \textbf{0.0387} & \textbf{0.0271} / \textbf{0.0312} & \textbf{0.0355} / \textbf{0.0413} \\
    \hline
  \end{tabular*}
  \caption{Performance on PPE Correctness subsets (dense models)}
  \label{tab:dense_ppe_subsets}
\end{table*}

\begin{table*}[h]
  \centering
  \footnotesize
  \begin{tabular}{lcc}
    \hline
    \textbf{Method} & \textbf{LLAMA Scout} & \textbf{GPT OSS 20B} \\
    \hline
    \multicolumn{3}{c}{\textbf{PPE Correctness Gpqa}} \\
    \hline
    Verbalized & 0.2631 / 0.2452 & 0.1505 / 0.1534 \\
    Consistency & 0.2684 / 0.2448 & \textbf{0.1084} / 0.1241 \\
    Majority & 0.2714 / 0.2443 & 0.1110 / \textbf{0.1176} \\
    Probe & \textbf{0.0652} / \textbf{0.0730} & 0.1301 / 0.1582 \\
    \hline
    \multicolumn{3}{c}{\textbf{PPE Correctness Ifeval}} \\
    \hline
    Verbalized & 0.1168 / 0.0904 & 0.1584 / 0.1627 \\
    Consistency & 0.3413 / 0.3186 & \textbf{0.0514} / 0.0625 \\
    Majority & 0.3410 / 0.3112 & 0.0517 / \textbf{0.0558} \\
    Probe & \textbf{0.0673} / \textbf{0.0803} & 0.0657 / 0.1049 \\
    \hline
    \multicolumn{3}{c}{\textbf{PPE Correctness Math}} \\
    \hline
    Verbalized & 0.2546 / 0.2505 & \textbf{0.0091} / \textbf{0.0122} \\
    Consistency & 0.2009 / 0.2066 & 0.0113 / 0.0205 \\
    Majority & 0.2061 / 0.2016 & 0.0115 / 0.0206 \\
    Probe & \textbf{0.0587} / \textbf{0.0653} & 0.0648 / 0.0790 \\
    \hline
    \multicolumn{3}{c}{\textbf{PPE Correctness Mbpp}} \\
    \hline
    Verbalized & \textbf{0.0289} / \textbf{0.0161} & 0.3545 / 0.3645 \\
    Consistency & 0.4737 / 0.4660 & 0.2765 / 0.2759 \\
    Majority & 0.4748 / 0.4669 & 0.2908 / 0.3009 \\
    Probe & 0.0453 / 0.0514 & \textbf{0.0472} / \textbf{0.0493} \\
    \hline
    \multicolumn{3}{c}{\textbf{PPE Correctness Mmlu}} \\
    \hline
    Verbalized & 0.2350 / 0.2186 & 0.0882 / 0.0894 \\
    Consistency & 0.2760 / 0.2559 & \textbf{0.0659} / 0.0754 \\
    Majority & 0.2750 / 0.2508 & 0.0691 / \textbf{0.0711} \\
    Probe & \textbf{0.0643} / \textbf{0.0684} & 0.0689 / 0.1193 \\
    \hline
  \end{tabular}
  \caption{Performance on PPE Correctness subsets (MOE models)}
  \label{tab:moe_ppe_subsets}
\end{table*}

\FloatBarrier
\subsection{Accuracy of Models on various Datasets}

We study calibration on models with varying difficulty for the models. RewardBench and PPE Math are the best performing across models.

PPE MBPP(Coding) and and PPE IfEval and PPE Preference are among the harder datasets across models.

\label{app:accuracy}
\begin{table*}[h]
  \centering
  \footnotesize
  \begin{tabular*}{\textwidth}{@{\extracolsep{\fill}}l*{8}{c}}
    \hline
    \textbf{Model} & \textbf{JudgeBench} & \textbf{PPE} & \textbf{PPE} & \textbf{PPE} & \textbf{PPE} & \textbf{PPE} & \textbf{PPE} & \textbf{Reward} \\
    & & \textbf{Preference} & \textbf{GPQA} & \textbf{IFEval} & \textbf{MBPP} & \textbf{MMLU} & \textbf{Math} & \textbf{Bench} \\
    \hline
\textbf{Average} & 0.670 & 0.627 & 0.633 & 0.624 & 0.583 & 0.733 & 0.793 & 0.856 \\
& {\scriptsize ±0.108} & {\scriptsize ±0.051} & {\scriptsize ±0.078} & {\scriptsize ±0.074} & {\scriptsize ±0.073} & {\scriptsize ±0.119} & {\scriptsize ±0.157} & {\scriptsize ±0.117} \\
    \hline
    LLAMA-8B & 0.549 & 0.577 & 0.548 & 0.553 & 0.508 & 0.589 & 0.606 & 0.743 \\
    J1-LLAMA-8B & 0.549 & 0.605 & 0.560 & 0.545 & 0.537 & 0.654 & 0.716 & 0.844 \\
    \hline
    LLAMA-70B & 0.644 & 0.660 & 0.624 & 0.605 & 0.599 & 0.750 & 0.694 & 0.890 \\
    J1-LLAMA-70B & 0.669 & 0.665 & 0.656 & 0.675 & 0.651 & 0.794 & 0.869 & 0.944 \\
    \hline
    QWEN-32B & 0.797 & 0.672 & 0.723 & 0.684 & 0.673 & 0.849 & 0.956 & 0.938 \\
    J1-QWEN-32B & 0.776 & 0.664 & 0.707 & 0.695 & 0.666 & 0.854 & 0.952 & 0.943 \\
    \hline
    GPT-OSS-20B & 0.797 & 0.639 & 0.711 & 0.705 & 0.533 & 0.821 & 0.953 & 0.923 \\
    \hline
    LLAMA-SCOUT & 0.578 & 0.531 & 0.532 & 0.530 & 0.501 & 0.555 & 0.598 & 0.621 \\
  \end{tabular*}
  \caption{Verbalized method accuracy across different models and datasets. Values show mean accuracy with standard deviation across models. RewardBench is the easiest for the models overall followed by PPE Math, PPE MBPP and PPE Preference are the hardest.}
  \label{tab:verbalized_accuracy}
\end{table*}

\subsection{Training the probe}
\label{training_probe}
 We use the HuggingFace transformers library to train the probes. The probe architecture has one linear layer with input dimension equal to the judge's hidden state dimension and output dimension of 1 (this is essentially linear regression). We train on 4000 examples randomly selected from the PPE datasets, with 2000 from all of the PPE correctness datasets and 2000 from the PPE preference dataset. The rest of the data ($\approx$10K examples) is used for in-distribution evaluation. We use a learning rate of $10^{-4}$, weight decay of $0.01$, batch size of 4, and train for 10 epochs. We experiment with two loss functions, mean squared error (MSE) or Brier Score Loss and Focal loss(FL), which is a reweighted binary cross entropy. The formulations of the two loss functions where $p$ is the prediction and $y$ is the label are as follows:
$$
\mathrm{MSE}(p, y) = (y - p)^2
$$
$$
\mathrm{FL}(p, y) = -\alpha \, y \, (1 - p)^\gamma \log(p) 
$$
$$ - (1 - \alpha) (1 - y) p^\gamma \log(1 - p)
$$
We take the probe's logits as the confidence scores, after thresholding them to be between 0 and 1.

\subsection{Layer Ablation Study}

\label{app:layer}
\begin{figure*}[htbp]
    \centering
    \includegraphics[width=0.8\textwidth]{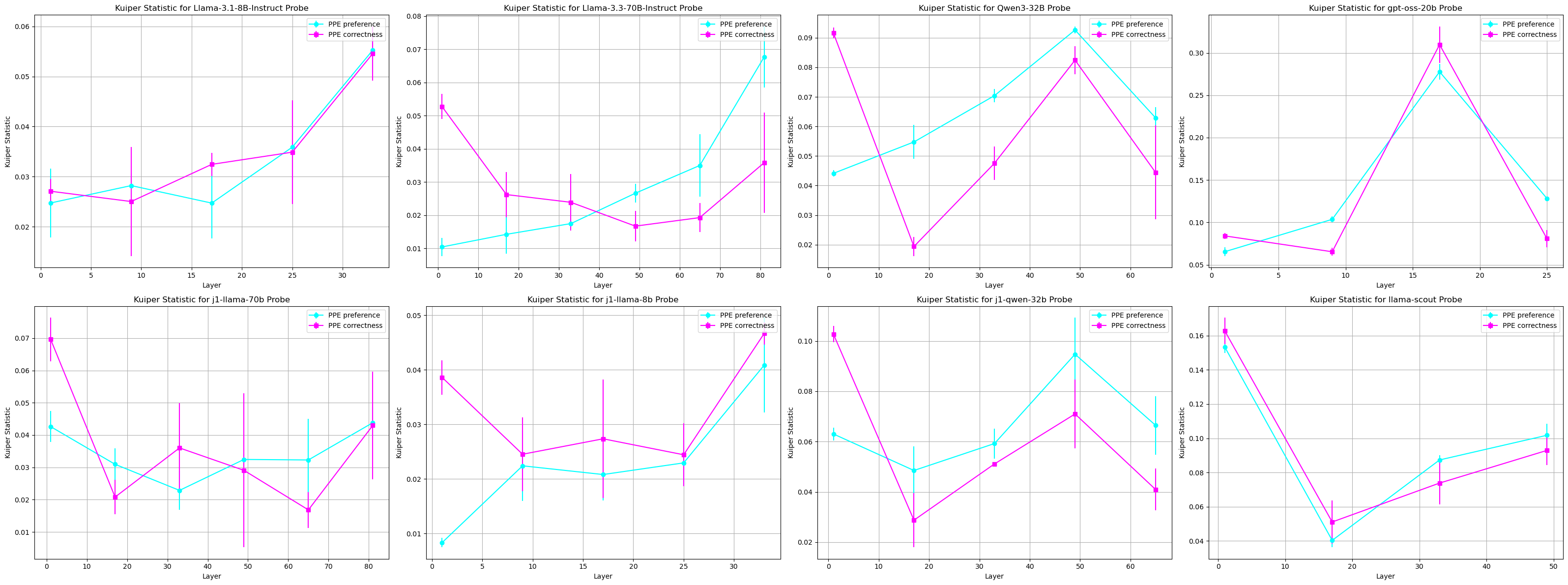}
    \caption{\textbf{Probes trained on the middle layers perform best.} Probe performance by transformer layer. Probes perform better when trained on middle layers, with performance typically peaking around layers 16-64 depending on model size.}
    \label{fig:layer_ablation_appendix}
\end{figure*}

We experiment with training the probes on different hidden state layers to determine the optimal layer for uncertainty estimation. We find that middle layers tend to perform better, and use this to inform our ultimate choice of which layers to report results on. Specifically, we choose layer 8 for GPT OSS 20B, layer 16 for the 8B and 32B models and LLaMA Scout, and layer 32 for the 70B models. Figure~\ref{fig:layer_ablation_appendix} shows the probe performance across different layers.

\FloatBarrier
\subsection{Loss Function Ablation}
\label{app:focal_loss_ablation}
We compare different loss functions for probe training, including focal loss with various hyperparameters $\alpha$ and $\gamma$, against MSE loss. Table~\ref{tab:loss_ablation} presents the results across different datasets and models.

Note focal loss with $\gamma=0$ is the binary-cross entropy loss.

\begin{table*}[htbp]
  \centering
  \footnotesize
  \begin{tabular}{lccc}
     \hline
    \textbf{Loss Function} & \textbf{LLaMA 70B} & \textbf{QWEN 32B} & \textbf{GPT OSS 20B} \\
    \hline
    \multicolumn{4}{c}{\textbf{PPE PREFERENCE (Kuiper / ECE)}} \\
    \hline
    Focal: $\alpha=0.25$, $\gamma=10$ & 0.5528 / 0.0986 & 0.1875 / 0.1266 & \textbf{0.1441} / \textbf{0.1138} \\
    Focal: $\alpha=0.5$, $\gamma=0$ & 0.1177 / 0.1644 & 0.1900 / 0.2111 & 0.2384 / 0.1869 \\
    Focal: $\alpha=0.75$, $\gamma=10$ & 0.2600 / 0.2522 & 0.2431 / 0.2402 & 0.2669 / 0.2452 \\
    Focal: $\alpha=0.75$, $\gamma=20$ & 0.0421 / 0.0629 & 0.1339 / 0.1629 & 0.2439 / 0.2326 \\
    MSE & \textbf{0.0303} / \textbf{0.0367} & \textbf{0.0765} / \textbf{0.1031} & 0.1676 / 0.1567 \\
    \hline
    \multicolumn{4}{c}{\textbf{PPE CORRECTNESS (Kuiper / ECE)}} \\
    \hline
    Focal: $\alpha=0.25$, $\gamma=10$ & 0.5865 / 0.2651 & 0.1433 / 0.1988 & \textbf{0.0738} / \textbf{0.1252} \\
    Focal: $\alpha=0.5$, $\gamma=0$ & 0.1364 / 0.1504 & 0.1221 / 0.1275 & 0.1270 / 0.1291 \\
    Focal: $\alpha=0.75$, $\gamma=10$ & 0.2234 / 0.2236 & 0.1380 / 0.1358 & 0.1440 / 0.1398 \\
    Focal: $\alpha=0.75$, $\gamma=20$ & 0.0500 / 0.0513 & 0.0821 / 0.0940 & 0.1365 / 0.1378 \\
    MSE & \textbf{0.0323} / \textbf{0.0367} & \textbf{0.0559} / \textbf{0.0814} & 0.0999 / 0.1325 \\
    \hline
    \multicolumn{4}{c}{\textbf{JUDGEBENCH (Kuiper / ECE)}} \\
    \hline
    Focal: $\alpha=0.25$, $\gamma=10$ & 0.5991 / 0.1400 & 0.1649 / 0.2177 & 0.1154 / 0.1671 \\
    Focal: $\alpha=0.5$, $\gamma=0$ & 0.1904 / 0.2176 & 0.1551 / 0.1526 & 0.1602 / 0.1619 \\
    Focal: $\alpha=0.75$, $\gamma=10$ & 0.2907 / 0.2823 & 0.1891 / 0.1936 & 0.1677 / 0.1688 \\
    Focal: $\alpha=0.75$, $\gamma=20$ & 0.0737 / 0.0967 & 0.1165 / 0.1220 & 0.1603 / 0.1639 \\
    MSE & \textbf{0.0546} / \textbf{0.0618} & \textbf{0.0682} / \textbf{0.0837} & \textbf{0.0986} / \textbf{0.1457} \\
    \hline
    \multicolumn{4}{c}{\textbf{REWARD BENCH (Kuiper / ECE)}} \\
    \hline
    Focal: $\alpha=0.25$, $\gamma=10$ & 0.6005 / 0.4541 & 0.1746 / 0.2379 & 0.1452 / 0.1992 \\
    Focal: $\alpha=0.5$, $\gamma=0$ & 0.0365 / 0.0652 & \textbf{0.0316} / 0.0800 & \textbf{0.0361} / 0.0962 \\
    Focal: $\alpha=0.75$, $\gamma=10$ & 0.0828 / 0.0852 & 0.0409 / \textbf{0.0468} & 0.0446 / \textbf{0.0742} \\
    Focal: $\alpha=0.75$, $\gamma=20$ & \textbf{0.0348} / \textbf{0.0443} & 0.0367 / 0.0700 & 0.0456 / 0.0944 \\
    MSE & 0.1165 / 0.1198 & 0.1232 / 0.1475 & 0.1427 / 0.1914 \\
    \hline
  \end{tabular}
  \caption{Ablations with different hyperparameters for focal loss. Note focal loss is equal to binary cross entropy loss at $\gamma=0$}
  \label{tab:loss_ablation}
\end{table*}

\subsection{Reliability Diagrams for all models on JudgeBench}
\label{app:reliability_diagrams}

We notice that verbalized confidence is generally overconfident across models. Multi-generation methods like majority or consistency are overconfident for the LLaMA family and the J1 LLaMa variants both across dense and MoE models. Similarly, for GPT-OSS-20B, models tend to get overconfident with multi-generation methods. However, for Qwen 32B and J1-Qwen 32B, consistency and majority methods make the model underconfident. 

\begin{figure}[htbp]
    \centering
    \begin{subfigure}{\textwidth}
        \centering
        \includegraphics[width=0.8\textwidth]{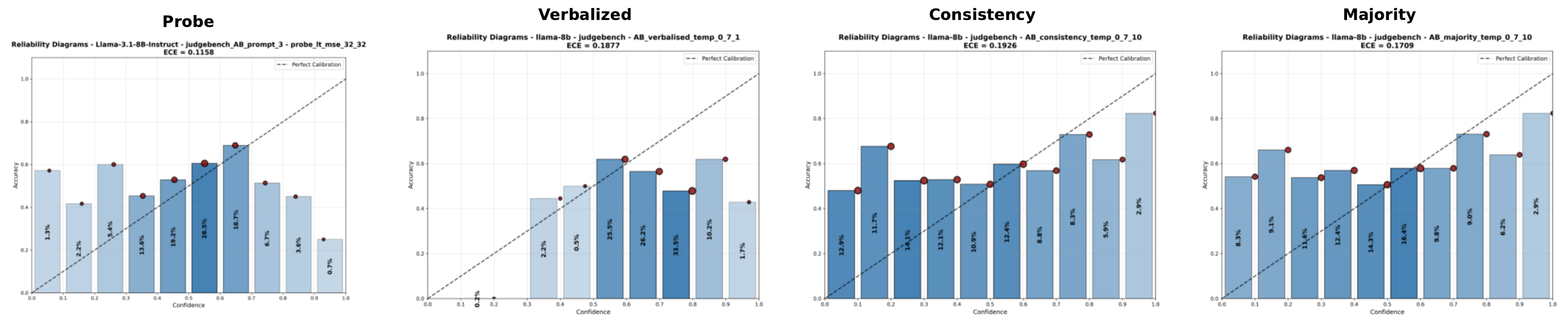}
        \caption{LLAMA 8B}
    \end{subfigure}
    \vspace{0.5em} 
    
    \begin{subfigure}{\textwidth}
        \centering
        \includegraphics[width=0.8\textwidth]{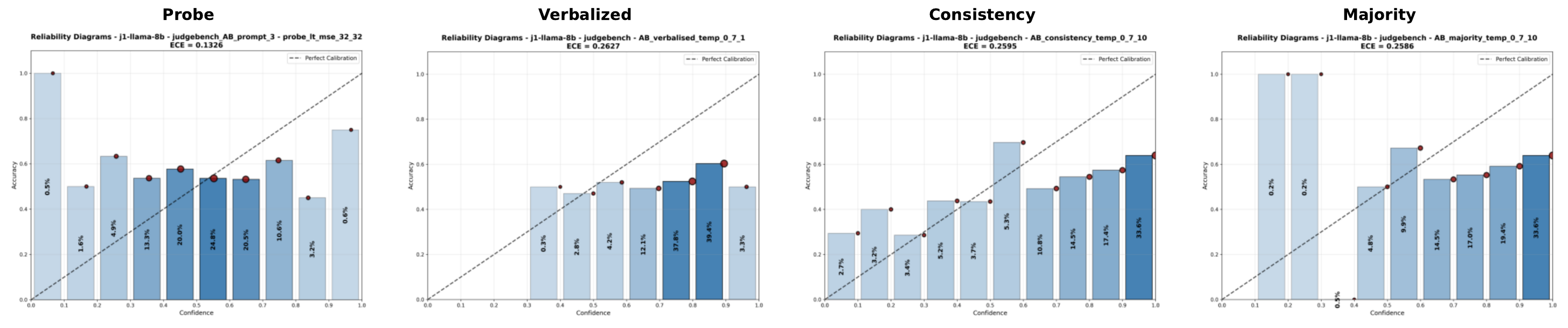}
        \caption{J1 LLAMA 8B}
    \end{subfigure}

    \begin{subfigure}{\textwidth}
        \centering
        \includegraphics[width=0.8\textwidth]{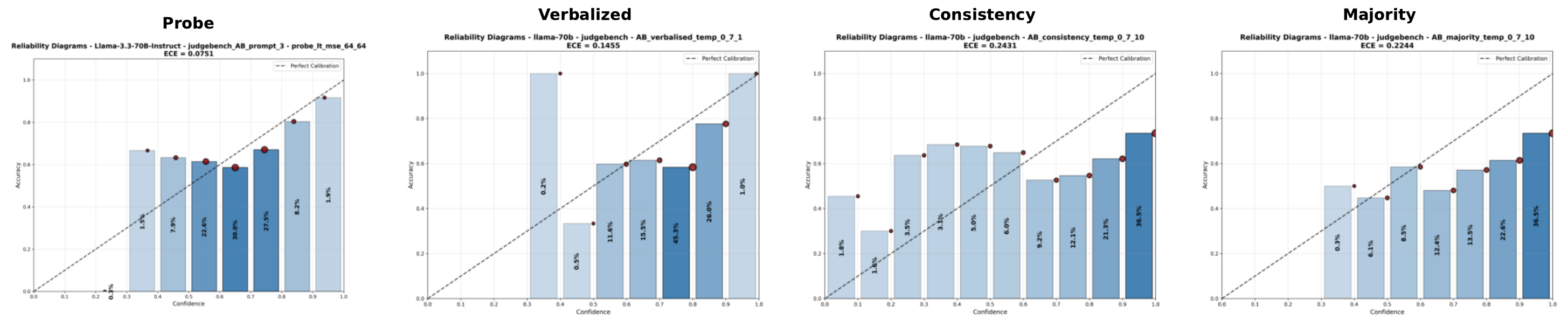}
        \caption{LLAMA 70B}
    \end{subfigure}

    \begin{subfigure}{\textwidth}
        \centering
        \includegraphics[width=0.8\textwidth]{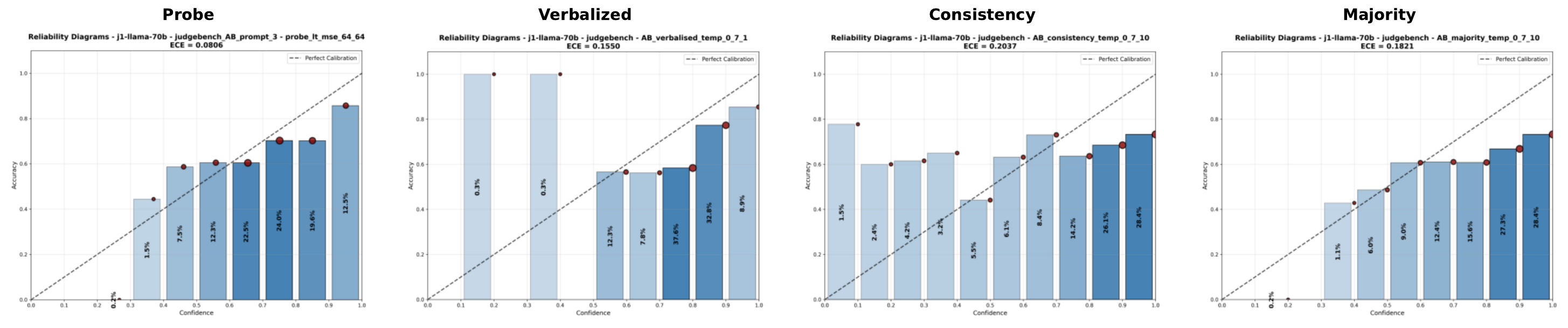}
        \caption{J1 LLAMA 70B}
    \end{subfigure}

    \begin{subfigure}{\textwidth}
        \centering
        \includegraphics[width=0.8\textwidth]{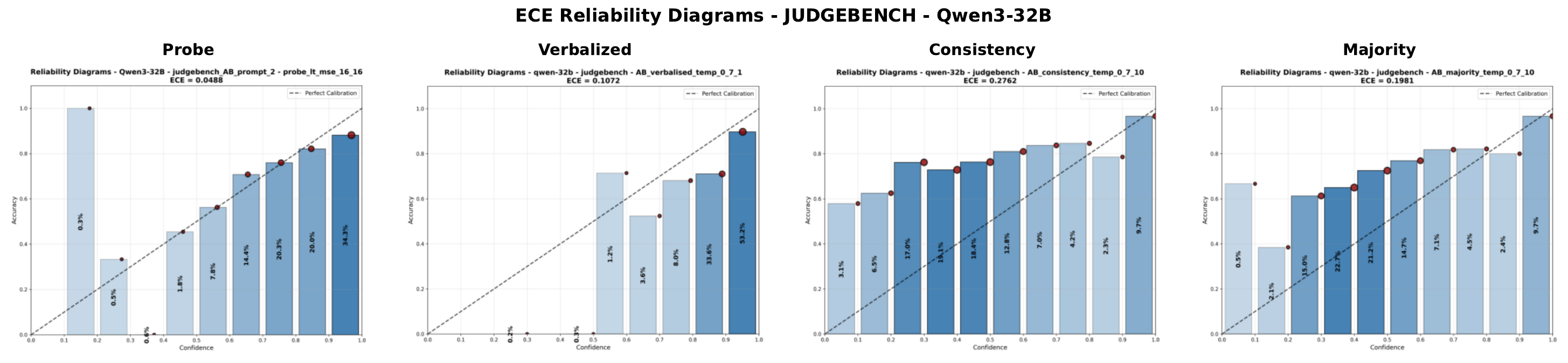}
        \caption{QWEN 32B}
    \end{subfigure}

    \begin{subfigure}{\textwidth}
        \centering
        \includegraphics[width=0.8\textwidth]{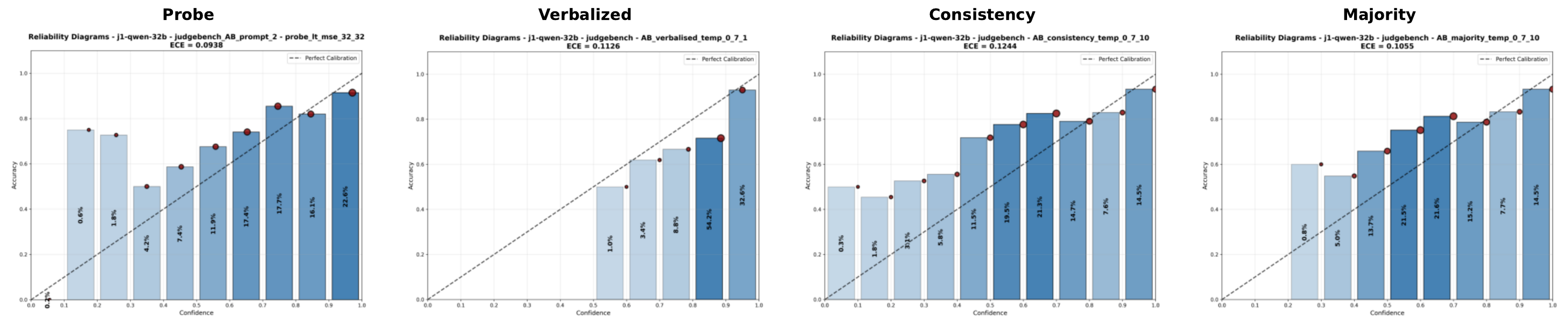}
        \caption{J1 QWEN 32B}
    \end{subfigure}

    \caption{Reliability plots for Dense Models}
    \label{fig:dense_model_rp}
\end{figure}

\begin{figure}[htbp]
    \centering
   \begin{subfigure}{\textwidth}
        \centering
        \includegraphics[width=0.8\textwidth]{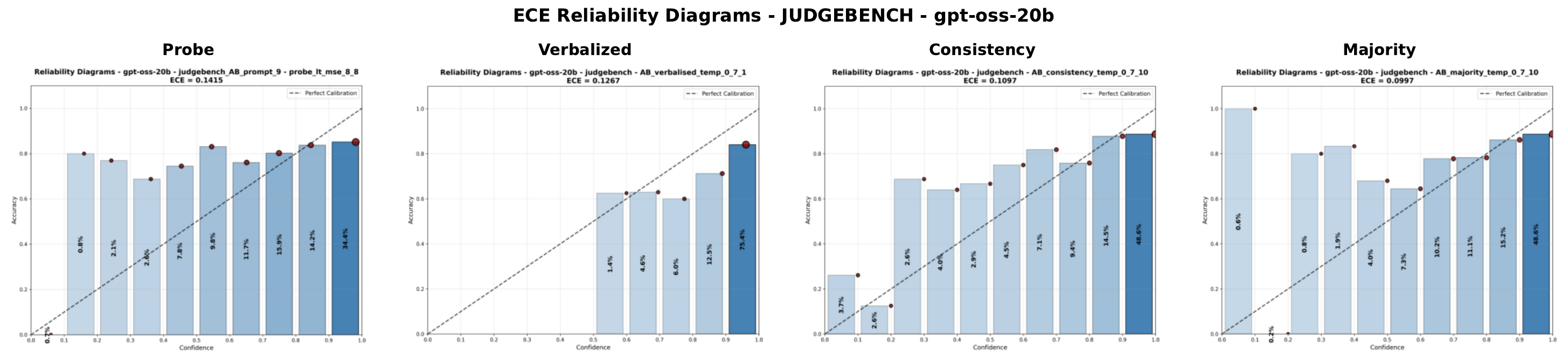}
        \caption{GPT OSS 20B}
    \end{subfigure}

    \begin{subfigure}{\textwidth}
        \centering
        \includegraphics[width=0.8\textwidth]{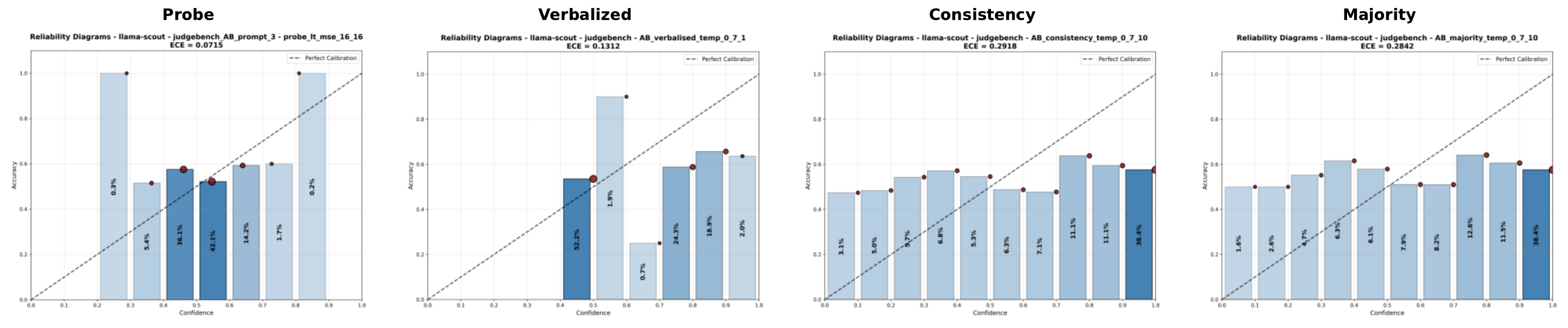}
        \caption{LLAMA SCOUT}
    \end{subfigure}
    \caption{Reliability plots for MoE Models}
    \label{fig:moe_model_rp}
\end{figure}

\FloatBarrier
\subsection{Temperature and Calibration Performance}
\label{temp_calib_perf}
We investigate the relationship between sampling temperature and calibration performance. Figure~\ref{fig:temp} shows how ECE and Kuiper statistics vary with temperature settings across different models.

\begin{figure*}[!htbp]
    \centering
    \includegraphics[width=0.8\linewidth]{figures/calibration_vs_temperature.pdf}
    \caption{ECE and Kuiper statistics as a function of temperature. We observe the lowest ECE and Kuiper values for most models at temperature 0.7, with minimal gains beyond this point. Performance is worst at temperature 0, improves at 0.7, and begins to decrease as temperature increases further.}
    \label{fig:temp} 
\end{figure*}

\subsection{Number of Runs and Consistency-Based Calibration}
\label{num_runs_calib_perf}
\begin{figure*}[!htbp]
    \centering
    \includegraphics[width=0.8\linewidth]{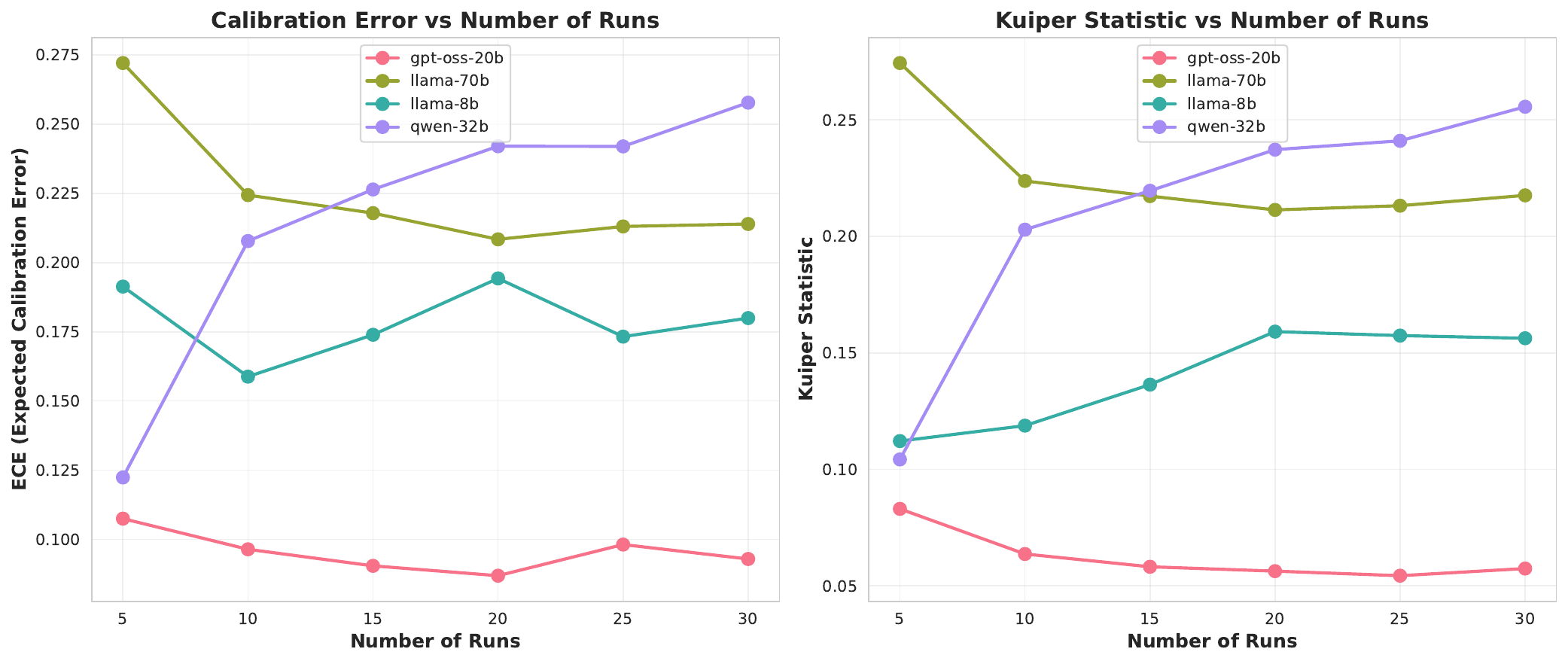}
    \caption{ECE and Kuiper statistics as a function of the number of consistency sampling runs. We observe the lowest ECE and Kuiper values for most models at 10 runs, with minimal gains beyond this point. Models tend to become overconfident as the number of runs increases.}
    \label{fig:num_runs}
\end{figure*}

We examine how the number of consistency sampling runs affects calibration performance. Figure~\ref{fig:num_runs} demonstrates the relationship between the number of runs and calibration metrics.

\subsection{Prompts used for Multigeneration (Consistency, Majority) baselines}

\begin{tcolorbox}[
  colback=gray!5,       
  colframe=gray!70!black, 
  title=Pairwise Judge with Verdicts
]
\label{prompt:verdict}
\textbf{Prompt:}

You are given a user question and two responses from two AI assistants. Your task is to act as an impartial judge and evaluate which response better follows the user’s instructions and provides a higher-quality answer.

First, provide your reasoning within \textless think\textgreater{} and \textless /think\textgreater{} tags. This should include your evaluation criteria for a high-quality response, a detailed comparison of the two responses, and when helpful, a reference answer as part of your evaluation. Be explicit in your thought process, referencing your criteria and explaining how each response aligns with or deviates from them.

If you are not entirely sure of the correctness of your verdict, you should verbally express your uncertainty in your thought process.

Avoid any position biases and ensure that the order in which the responses were presented does not influence your decision. Do not allow the length of the responses to influence your evaluation. Do not favor certain names of the assistants. Be as objective as possible.

Finally, provide your verdict within \textless answer\textgreater{} and \textless /answer\textgreater{} tags, strictly following this format:
- \textless answer\textgreater{} [[A]] \textless /answer\textgreater{} if Assistant A is better
- \textless answer\textgreater{} [[B]] \textless /answer\textgreater{} if Assistant B is better

Below are the user’s question and the two responses:

\bigskip
User Question]  
\{input\}

[The Start of Assistant A's Answer]  

\{response\_1\}  

[The End of Assistant A's Answer]

[The Start of Assistant B's Answer]

\{response\_2\}  

[The End of Assistant B's Answer]

\end{tcolorbox}

\begin{tcolorbox}[
  colback=gray!5,       
  colframe=gray!70!black, 
  title=Pairwise Judge with Scores,
]
\label{prompt:scores}
\textbf{Prompt:}
You are given a user question and two responses from two AI assistants. Your task is to act as an impartial judge and evaluate which response better follows the user’s instructions and provides a higher-quality answer.

First, provide your reasoning within \textless think\textgreater{} and \textless /think\textgreater{} tags. This should include your evaluation criteria for a high-quality response, a detailed comparison of the two responses, and when helpful, a reference answer as part of your evaluation. Be explicit in your thought process, referencing your criteria and explaining how each response aligns with or deviates from them.

If you are not entirely sure of the correctness of your verdict, you should verbally express your uncertainty in your thought process.

Avoid any position biases and ensure that the order in which the responses were presented does not influence your decision. Do not allow the length of the responses to influence your evaluation. Do not favor certain names of the assistants. Be as objective as possible.

Finally, assign the assistant’s response a score from 0 to 10, using either an integer or a decimal with up to 0.1 precision, with a higher score indicating a higher-quality response that better satisfies the criteria. Enclose the scores within the tags \textless score\_A\textgreater{} \textless /score\_A\textgreater{}, and \textless score\_B\textgreater{} \textless /score\_B\textgreater{}.

Format your output like this: \\
\textless think\textgreater{} your\_thinking\_process \textless /think\textgreater{} \\
\textless score\_A\textgreater{} your\_score\_a \textless /score\_A\textgreater{} \ \textless score\_B\textgreater{} your\_score\_b \textless /score\_B\textgreater{}

Below are the user’s question and the two responses:

\bigskip
[User Question]

\{input\}

[The Start of Assistant A's Answer]  

\{response\_1\}  

[The End of Assistant A's Answer]

[The Start of Assistant B's Answer]

\{response\_2\}  

[The End of Assistant B's Answer]

\end{tcolorbox}

\begin{tcolorbox}[
  colback=gray!5,
  colframe=gray!70!black,
  title=Pairwise Judge with Likert Score -- LM Arena Hard Prompt,
]
\label{prompt:likert}

Please act as an impartial judge and evaluate the quality of the responses provided by two AI assistants to the user prompt displayed below. You will be given Assistant A's answer and Assistant B's answer. Your job is to evaluate which assistant's answer is better.

Begin your evaluation by generating your own answer to the prompt. You must provide your answers before judging any answers.

When evaluating the assistants' answers, compare both assistants' answers with your answer. You must identify and correct any mistakes or inaccurate information.

Then consider if the assistants' answers are helpful, relevant, and concise. Helpful means the answer correctly responds to the prompt or follows the instructions. Note that when a user prompt has any ambiguity or more than one interpretation, it is more helpful and appropriate to ask for clarifications or more information from the user than providing an answer based on assumptions. Relevant means all parts of the response closely connect or are appropriate to what is being asked. Concise means the response is clear and not verbose or excessive.

Then consider the creativity and novelty of the assistants' answers when needed. Finally, identify any missing important information in the assistants' answers that would be beneficial to include when responding to the user prompt.

Verbalize any uncertainty in your thought process when you're unsure of your conclusion.

If your verdict is not based on absolute certainty, be sure to articulate that hesitation in your analysis.

After providing your explanation, you must output only one of the following choices as your final verdict with a label:

1. Assistant A is significantly better: [[A>>B]] \\
2. Assistant A is slightly better: [[A>B]] \\
3. Tie, relatively the same: [[A=B]] \\
4. Assistant B is slightly better: [[B>A]] \\
5. Assistant B is significantly better: [[B>>A]]

Example output: ``My final verdict is tie: [[A=B]].''

\bigskip
[User Question]

\{input\}

[The Start of Assistant A's Answer]  

\{response\_1\}  

[The End of Assistant A's Answer]

[The Start of Assistant B's Answer]

\{response\_2\}  

[The End of Assistant B's Answer]

\end{tcolorbox}

\subsection{Prompts for Verbalized confidence}

\begin{tcolorbox}[
  colback=gray!5,       
  colframe=gray!70!black, 
  title=Pairwise Judge with Verdicts,
]
\label{prompt:verdict}
\textbf{Prompt:}

You are given a user question and two responses from two AI assistants. Your task is to act as an impartial judge and evaluate which response better follows the user’s instructions and provides a higher-quality answer.

First, provide your reasoning within \textless think\textgreater{} and \textless /think\textgreater{} tags. This should include your evaluation criteria for a high-quality response, a detailed comparison of the two responses, and when helpful, a reference answer as part of your evaluation. Be explicit in your thought process, referencing your criteria and explaining how each response aligns with or deviates from them.

If you are not entirely sure of the correctness of your verdict, you should verbally express your uncertainty in your thought process.

Avoid any position biases and ensure that the order in which the responses were presented does not influence your decision. Do not allow the length of the responses to influence your evaluation. Do not favor certain names of the assistants. Be as objective as possible.

Finally, provide your verdict within \textless answer\textgreater{} and \textless /answer\textgreater{} tags, strictly following this format:
- \textless answer\textgreater{} [[A]] \textless /answer\textgreater{} if Assistant A is better
- \textless answer\textgreater{} [[B]] \textless /answer\textgreater{} if Assistant B is better

\bigskip
Also, provide a confidence score (0-100) for your judgment, representing how likely your answer is to be correct, enclosed within <confidence> and </confidence> tags.

- <confidence> 50 </confidence>

Below are the user’s question and the two responses:

\bigskip
User Question]  
\{input\}

[The Start of Assistant A's Answer]  

\{response\_1\}  

[The End of Assistant A's Answer]

[The Start of Assistant B's Answer]

\{response\_2\}  

[The End of Assistant B's Answer]

\end{tcolorbox}

\begin{tcolorbox}[
  colback=gray!5,       
  colframe=gray!70!black, 
  title=Pairwise Judge with Scores,
]
\label{prompt:scores}
\textbf{Prompt:}
You are given a user question and two responses from two AI assistants. Your task is to act as an impartial judge and evaluate which response better follows the user’s instructions and provides a higher-quality answer.

First, provide your reasoning within \textless think\textgreater{} and \textless /think\textgreater{} tags. This should include your evaluation criteria for a high-quality response, a detailed comparison of the two responses, and when helpful, a reference answer as part of your evaluation. Be explicit in your thought process, referencing your criteria and explaining how each response aligns with or deviates from them.

If you are not entirely sure of the correctness of your verdict, you should verbally express your uncertainty in your thought process.

Avoid any position biases and ensure that the order in which the responses were presented does not influence your decision. Do not allow the length of the responses to influence your evaluation. Do not favor certain names of the assistants. Be as objective as possible.

Finally, assign the assistant’s response a score from 0 to 10, using either an integer or a decimal with up to 0.1 precision, with a higher score indicating a higher-quality response that better satisfies the criteria. Enclose the scores within the tags \textless score\_A\textgreater{} \textless /score\_A\textgreater{}, and \textless score\_B\textgreater{} \textless /score\_B\textgreater{}.

Format your output like this: \\
\textless think\textgreater{} your\_thinking\_process \textless /think\textgreater{} \\
\textless score\_A\textgreater{} your\_score\_a \textless /score\_A\textgreater{} \ \textless score\_B\textgreater{} your\_score\_b \textless /score\_B\textgreater{}

\bigskip

Also, provide a confidence score (0-100) for your judgment, representing how likely your answer is to be correct, enclosed within <confidence> and </confidence> tags.

- <confidence> 50 </confidence>

Below are the user’s question and the two responses:

\bigskip
[User Question]

\{input\}

[The Start of Assistant A's Answer]  

\{response\_1\}  

[The End of Assistant A's Answer]

[The Start of Assistant B's Answer]

\{response\_2\}  

[The End of Assistant B's Answer]

\end{tcolorbox}

\begin{tcolorbox}[
  colback=gray!5,       
  colframe=gray!70!black,
  title=Pairwise Judge with Likert Score -- Chatbot Arena Hard Prompt,
]
\label{prompt:likert}

Please act as an impartial judge and evaluate the quality of the responses provided by two AI assistants to the user prompt displayed below. You will be given Assistant A's answer and Assistant B's answer. Your job is to evaluate which assistant's answer is better.

Begin your evaluation by generating your own answer to the prompt. You must provide your answers before judging any answers.

When evaluating the assistants' answers, compare both assistants' answers with your answer. You must identify and correct any mistakes or inaccurate information.

Then consider if the assistants' answers are helpful, relevant, and concise. Helpful means the answer correctly responds to the prompt or follows the instructions. Note that when a user prompt has any ambiguity or more than one interpretation, it is more helpful and appropriate to ask for clarifications or more information from the user than providing an answer based on assumptions. Relevant means all parts of the response closely connect or are appropriate to what is being asked. Concise means the response is clear and not verbose or excessive.

Then consider the creativity and novelty of the assistants' answers when needed. Finally, identify any missing important information in the assistants' answers that would be beneficial to include when responding to the user prompt.

\bigskip
Verbalize any uncertainty in your thought process when you're unsure of your conclusion.

If your verdict is not based on absolute certainty, be sure to articulate that hesitation in your analysis.

Your judgment must also include a confidence score (0–100) reflecting its likelihood of being accurate.

\bigskip
After providing your explanation, you must output only one of the following choices as your final verdict with a label:

1. Assistant A is significantly better: [[A>>B]] \\
2. Assistant A is slightly better: [[A>B]] \\
3. Tie, relatively the same: [[A=B]] \\
4. Assistant B is slightly better: [[B>A]] \\
5. Assistant B is significantly better: [[B>>A]]

\bigskip
Example output: "My final verdict is tie: [[A=B]]. Confidence: 0.85".

\bigskip
[User Question]

\{input\}

[The Start of Assistant A's Answer]  

\{response\_1\}  

[The End of Assistant A's Answer]

[The Start of Assistant B's Answer]

\{response\_2\}  

[The End of Assistant B's Answer]

\end{tcolorbox}

\subsection{Prompt ablations for verbalized confidence}
\label{verbalised_ablations}

We also conduct ablation studies to determine which approach yields the most effective verbalized confidence from the models. Specifically, we experiment with permitting expressions of uncertainty within the chain-of-thought reasoning, as opposed to the default chain-of-thought approach. Additionally, we test different confidence scales, including both the 0–1 and 0–100 ranges.

\subsubsection{Varying confidence ranges}

\begin{tcolorbox}[  colback=gray!5,       
  colframe=gray!70!black, title=Prompt 1: Range 0-1]
"Also, provide a confidence score (0-1) for your judgment, representing how confident you are about your answer, enclosed within <confidence> and </confidence> tags."

\textbf{Example output:} \texttt{<confidence> 0.3 </confidence>}
\end{tcolorbox}

\begin{tcolorbox}[  colback=gray!5,       
  colframe=gray!70!black,title=Prompt 2: Range 0-100]
"Also, provide a confidence score (0-100) for your judgment, representing how confident you are about your answer, enclosed within <confidence> and </confidence> tags."

\textbf{Example output:} \texttt{<confidence> 30 </confidence>}
\end{tcolorbox}

\subsubsection{Varying elicitation mode}
\begin{tcolorbox}[  colback=gray!5,       
  colframe=gray!70!black, title=Prompt 3: With uncertainty in reasoning]
"If you are not entirely sure of the correctness of your verdict, you should verbally express your uncertainty in your thought process. Also, provide a confidence score (0-100) for your judgment, representing how likely your answer is to be correct, enclosed within <confidence> and </confidence> tags."

\textbf{Example output:} \texttt{<confidence> 50 </confidence>}
\end{tcolorbox}

\begin{tcolorbox}[  colback=gray!5,       
  colframe=gray!70!black, title=Prompt 4: Without uncertainty in reasoning]
"Also, provide a confidence score (0-100) for your judgment, representing how confident you are about your answer, enclosed within <confidence> and </confidence> tags."

\textbf{Example output:} \texttt{<confidence> 30 </confidence>}
\end{tcolorbox}

\begin{table}[h!]
\begin{tabular}{l|cc|cc}
\hline
\textbf{Model} 
  & \multicolumn{2}{c|}{\textbf{With uncertainty in reasoning}} 
  & \multicolumn{2}{c}{\textbf{Without uncertainty in reasoning}} \\
{} 
  & \textbf{0-1} & \textbf{0-100} 
  & \textbf{0-1} & \textbf{0-100} \\
\hline
LLAMA 70b & 0.1483 & \textbf{0.14496} & 0.19916 & 0.17803 \\
J1 LLAMA 70b          & \textbf{0.15651} & 0.16230 & 0.18181 & 0.19956 \\
LLAMA 8b              & 0.18799 & \textbf{0.14251} & 0.26258 & 0.20280 \\
J1 LLAMA 8b           & 0.26245 & 0.18474 & 0.21141 & \textbf{0.17139} \\
\hline

\end{tabular}
\caption{Calibration (Kuiper, lower is better) for various prompts eliciting verbalized confidence}
\end{table}

\begin{figure}[!htbp]
    \centering
    \includegraphics[width=0.5\textwidth]{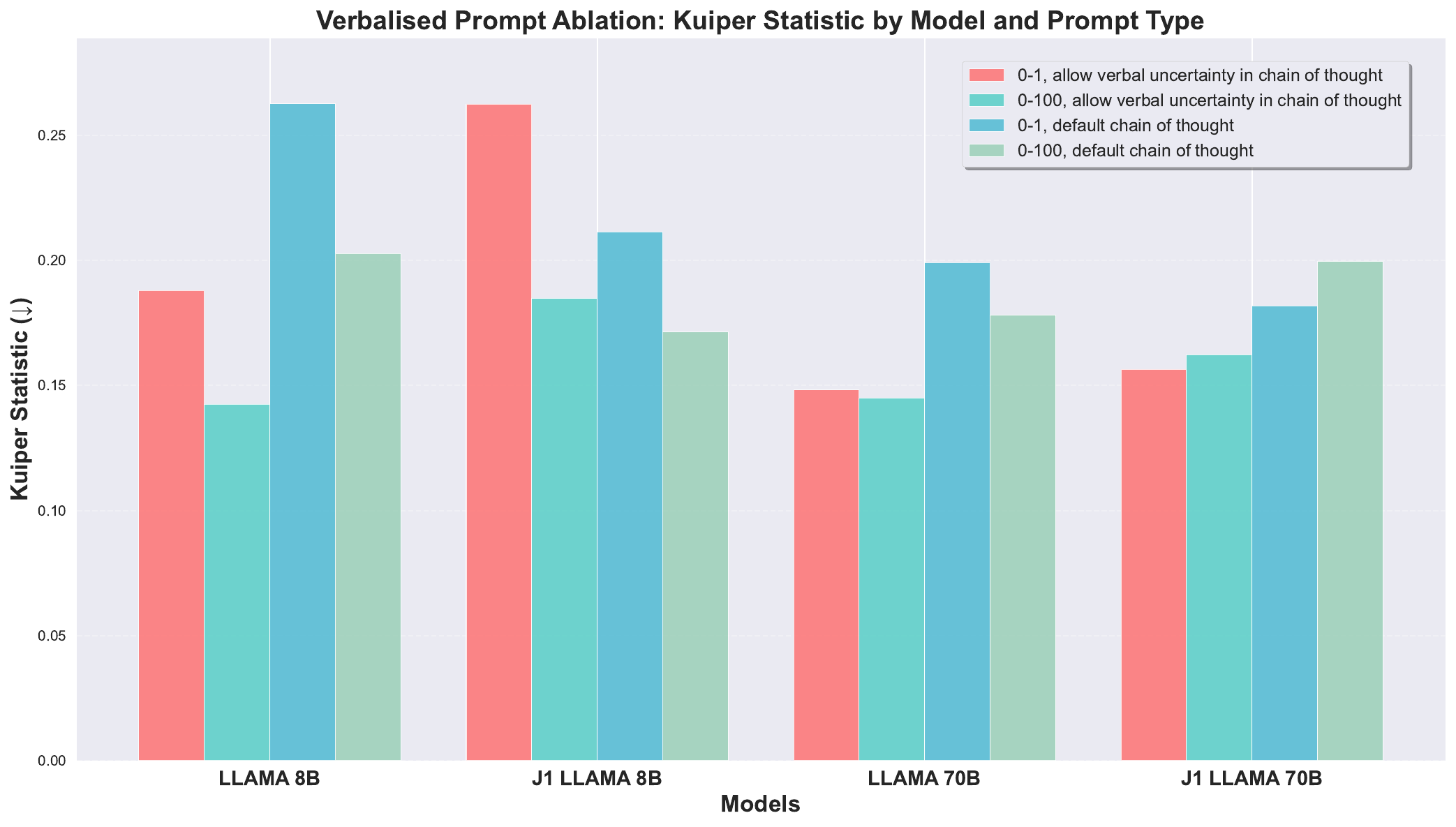}
    \caption{Best prompts for Verbalized confidence as measured by the Kuiper metric. Lower is better. 0-100 ranges and allowing uncertainty in reasoning work best.}
    \label{fig:verbalised_ablation}
\end{figure}

\end{document}